\documentclass[twocolumn,letterpaper]{IEEEAerospaceCLS}  

\usepackage[]{graphicx}    
\usepackage{gensymb}
\usepackage{url}
\usepackage{enumitem}

\setlist{leftmargin=6mm}

\newcommand{\ignore}[1]{}  

\begin{document}
\title{LunarNav: Crater-based Localization for Long-range Autonomous Lunar Rover Navigation}

\author{%
Shreyansh Daftry, Zhanlin Chen, Yang Cheng, Scott Tepsuporn, Shehryar Khattak and Larry Matthies\\ 
Jet Propulsion Laboratory, California Institute of Technology, USA\\
Pasadena, CA 91109\\
\texttt{Shreyansh.Daftry@jpl.nasa.gov}\\
\and 
Brian Coltin, Ussama Naal, Lanssie Mingyue Ma and Matthew Deans\\
NASA Ames Research Center\\
Mountain View, CA 94043\\
\texttt{brian.coltin@nasa.gov}\\
\thanks{\footnotesize 978-1-6654-9032-0/23/$\$31.00$ \copyright2023 IEEE}              
}

\maketitle

\thispagestyle{plain}
\pagestyle{plain}

\maketitle

\thispagestyle{plain}
\pagestyle{plain}

\begin{abstract}
The Artemis program requires robotic and crewed lunar rovers for resource prospecting and exploitation, construction and maintenance of facilities, and human exploration. These rovers must support navigation for 10s of kilometers (km) from base camps. A lunar science rover mission concept (“Endurance-A”) has been recommended by the new Decadal Survey as the highest priority medium-class mission of the Lunar Discovery and Exploration Program, and would be required to traverse approximately 2000 km in the South Pole-Aitkin (SPA) Basin, with individual drives of several kilometers between stops for downlink. These rover mission scenarios require functionality that provides onboard, autonomous, global position knowledge (“absolute localization”). However, planetary rovers have no onboard global localization capability to date; they have only used relative localization, by integrating combinations of wheel odometry, visual odometry, and inertial measurements during each drive to track position relative to the start of each drive. At the end of each drive, a “ground-in-the-loop” (GITL) interaction is used to get an absolute position update from human operators in a more global reference frame. As a result, autonomous rover drives are limited in distance so that accumulated relative navigation error does not risk the possibility of the rover driving into a “keep-out zone”; in practice, drive limits of a few hundred meters are to be expected.

In this work, we summarize recent developments from the LunarNav project, where we have developed algorithms and software to enable lunar rovers to estimate their global position and heading on the Moon with a goal performance of position error less than 5 meters (m) and heading error less than 3$^{\circ}$, 3$\sigma$, in sunlit areas. This new capability will eliminate the need for GITL interactions with human operators for lunar rover global position estimation, which will substantially increase operational productivity of lunar rovers and will reduce operations costs. This will be achieved autonomously onboard by detecting craters in the vicinity of the rover and matching them to a database of known craters mapped from orbit. The overall technical framework consists of three main elements: 1) crater detection, 2) crater matching, and 3) state estimation. In previous work, we developed crater detection algorithms for three different sensing modalities. This paper builds on that work, and focuses on the crater matching and state estimation aspects of the problem. In particular, we developed two algorithms for crater-based localization, and demonstrated them on datasets of both real and simulated lunar data, in representative environments. Our results suggest that rover localization with an error less than 5 m is highly probable during daytime operations.
\end{abstract} 

\tableofcontents

\section{Introduction}
Long-range, autonomous lunar rover mobility is expected to be part of human exploration and robotic science missions to the Moon in the next decade and beyond. Multiple whitepapers and mission concept studies \cite{psads2022,intrepid,elliott2020intrepid,heldman2022inspire,keane2022endurance} that addressed science that could be done with robotic lunar rovers were submitted to the recent Planetary Science and Astrobiology Decadal Survey (PSADS). The final report from PSADS \cite{national2022origins} recommended a long-range lunar rover mission concept called “Endurance-A" as the highest priority medium-class mission for the Moon in the next decade. This mission concept would traverse $\sim$2000 km in the South Pole Aitken (SPA) basin to collect $\sim$100 kg of samples, which would be returned to Earth by astronauts. NASA’s Artemis program for returning astronauts to the Moon includes a concept for a Lunar Terrain Vehicle (LTV) that would transport astronauts over distances of up to 20 km without having to stop to recharge batteries and would be capable of operation remotely when astronauts are not present \cite{ltv}. These rover concepts would drive quickly compared to prior rovers: up to 30 cm/s for robotic science rovers and probably much faster for an LTV. They would also cover several kilometers (km) per autonomous rover drive, compared to at most a few hundred meters per drive for prior robotic science rovers.

All of these lunar rover mission concepts would require knowledge of the rover’s position in a global reference frame. Since they would operate far beyond line-of-sight from a lander, they would need to obtain position knowledge independently from the lander need to obtain position knowledge from means other than navigation aiding from a lander. The accuracy needed for rover global position knowledge is debatable, but on the order of 10 m is desirable for confidence that a rover will not stray into known hazards during long periods of autonomous driving with no contact with human operators.

Planetary rovers to date have all used onboard dead reckoning for position estimation relative to the start of each day’s drive, with global position updates obtained with the aid of humans using rover images transmitted to Earth. This would not support these lunar rover mission concepts, because of unacceptably high requirements for communication with Earth and for support from human operators on Earth. Several methods of onboard position estimation have been studied, as reviewed in \cite{matthies2022lunar}: recognizing horizon landmarks, registering local elevation maps created onboard with elevation maps created from orbit, radio frequency navigation aiding from one or more orbiters, and celestial navigation using measured vectors to the Sun, the Earth, and the Moon’s core. These methods have a variety of limitations, including position error often on the order of 100 m or more, limited availability of sufficiently high-resolution elevation maps from orbit, and relatively long periods when orbiter(s) are not overhead to provide RF navigation aiding. 

An alternative approach is to use craters as landmarks, with the rover automatically detecting craters in its vicinity and matching them to known craters in a landmark database created from orbital imagery. This approach has the advantage that almost all sunlit areas of the Moon have been imaged from orbit with enough resolution ($\sim$ 0.5 m/pixel) to enable mapping crater landmarks as small as 5 to 10 m in diameter; moreover, the ShadowCam camera \cite{robinson2018shadowcam} on the Korean Pathfinder Lunar Orbiter is expected to map permanently shadowed regions with a resolution of $\sim$ 1.7 m/pixel, which could enable mapping crater landmarks with diameters of 10 to 20 m. This approach to lunar rover localization was introduced in \cite{matthies2022lunar}, which described and evaluated algorithms for detecting craters in the vicinity of a rover with point cloud data from onboard lidar or stereo cameras, and/or with neural net-based pattern recognition in monocular images. Here, we build on that work by presenting methods to match such onboard crater detections to craters in the landmark database and to update rover global position estimates with a sequence of such detections as the rover moves. A related paper focuses on crater-based localization in the dark and for permanently shadowed regions \cite{cauligi2023shadownav}

The rest of the paper is organized as follows: Section 2 discusses related work. Section 3 describes LunarNav's overall crater-based localization system concept. Section 4 describes data sets of real and synthetic images obtained for this work. Section 5 describes the technical approach and two algorithms developed for the crater-based localization. Results of quantitative performance evaluation for both algorithms using multiple sensing modalities are shown in Section 6. These results show this approach to be very promising for achieving the goal of rover localization on the lunar surface, with an error less than 5m during daytime operations. Section 7 summarizes the results and outlines planned future work.

\vspace{2mm}
\label{sec:intro}

\section{Related Work}
\vspace{2mm}
Methods for lunar rover localization were surveyed in \cite{matthies2022lunar}; here we give a brief recap of prior and an update on recent related work. As noted in \cite{matthies2022lunar}, applicable methods fall into several classes:

\begin{itemize}
    \setlength\itemsep{0.5em}
    \item Registering onboard image or local map data to orbital images or maps
    \item Using horizon features as landmarks
    \item Celestial measurements, potentially combined with other measurements, such as accurate gravity vector measurements
    \item Radiometric ranging from lunar orbit or Earth tracking stations
\end{itemize}

Transferring rover images to Earth for human operators to register them to orbital images has been the standard approach for Mars rover localization of Mars rovers \cite{parker2010geomorphic}; this can achieve accuracy on the scale of the resolution of the orbital images, which is sub-meter scale for Mars and the Moon. This is more difficult to do when the sun angle varies as much as it would in lunar rover missions, especially given the high-contrast shadows on the Moon. This motivates using craters as landmarks, because these can be detected reliability independently of sun angle. Algorithms to register local elevation maps created onboard to elevation maps created with orbital imagery are described in \cite{gaines2020self}. A limitation of this approach for the Moon is that it requires high resolution stereo imaging from orbit of the entire rover traverse, which is not currently available.

The accuracy of localization methods based on horizon features depends on the terrain relief, distance to the horizon features, and resolution of the associated elevation map; in favorable situations this can give errors $<$ 50 m \cite{chiodini2017mars}, but this is not always available. Simulations of celestial navigation methods show 1$\sigma$ errors on the order of 100 m \cite{yang2014simultaneous}. In principle, radio ranging between the rover and either a lunar orbiter or Earth tracking stations can give meter-scale position knowledge \cite{chelmins2009kalman}; however, the orbiter infrastructure needed for this is not in place and this would not provide instant position updates or continuous service.

Automatic crater detection in orbital images has been done to determine crater statistics for scientific purposes \cite{yang2020lunar} and to estimate the position of landers during descent to highly craters asteroids \cite{cheng2003optical}. Crater detection for lunar rover localization was introduced in \cite{matthies2022lunar}. Using landmarks like this for localization over long distances requires methods to reliably correspond craters detected onboard with craters in the landmark databased, despite errors in both onboard crater detection and the orbital landmark map. The large literature on “Simultaneous Localization and Mapping” (SLAM) is applicable to this problem, and includes methods such as particle filters and incremental batch optimization with crater detection \cite{thrun2002probabilistic,cornejo2020survey}. These methods currently appear to be the most promising approach to achieving lunar rover global localization with the required accuracy and update frequency.

\section{System Overview}
\vspace{2mm}
\begin{figure*}[!tbh]
    \centering
    \includegraphics[width=\linewidth]{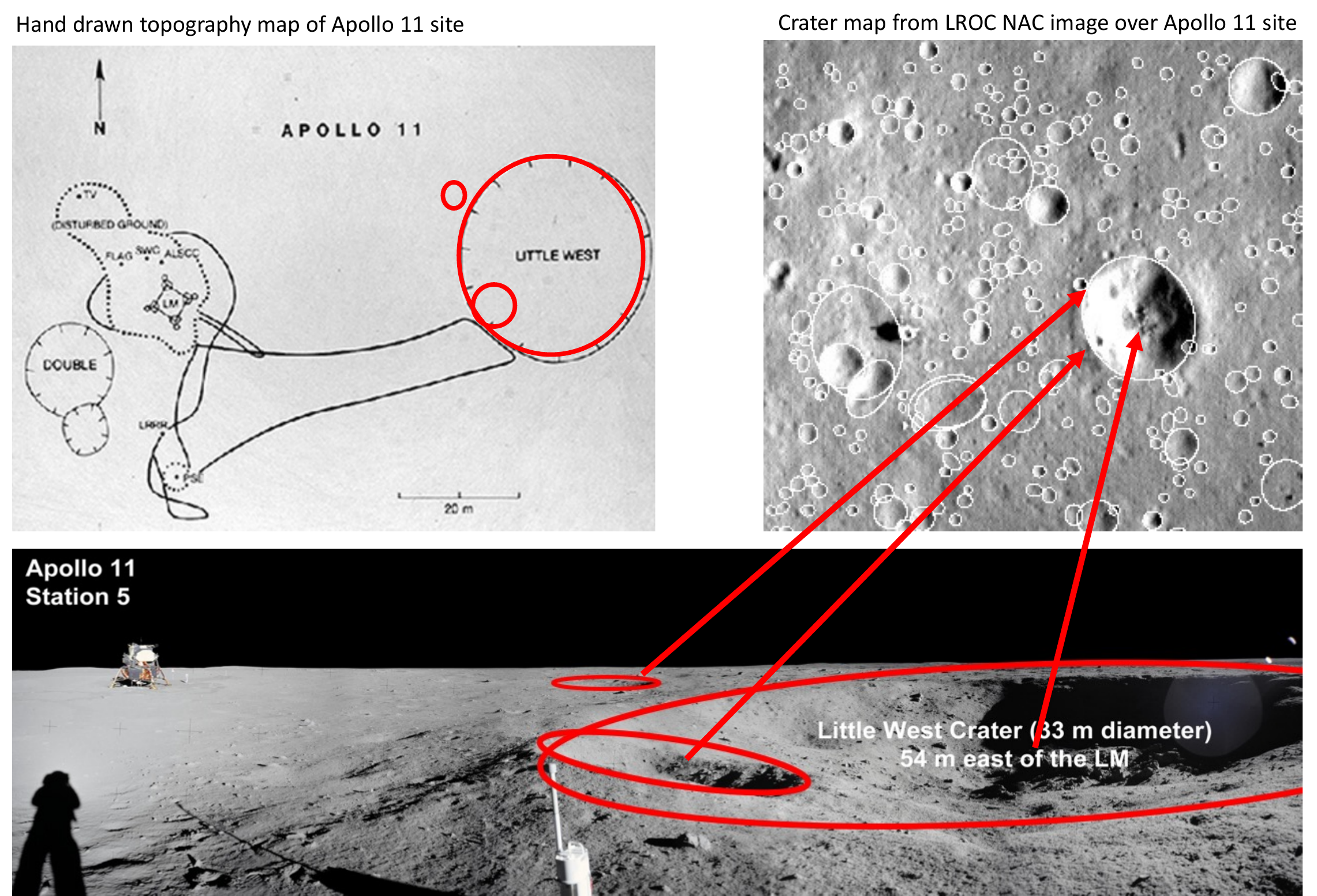}
    \caption{\bf{Illustration of LunarNav’s overall system concept to recognize craters from surface images and match to the crater map will help to localize rover globally}}
    \label{fig:system_concept}
\end{figure*}

\subsection{System Concept}
The LunarNav system concept requires creating a database of crater landmarks from orbital images, as shown in Figure \ref{fig:system_concept}. This database would contain the position, diameter, and estimated depth of each crater. Craters with diameters $> 5$ meters are mappable with LRO-NAC imagery, under most orbital imaging conditions. Even in the youngest terrain on the Moon, such craters occur with a frequency $> 10^3$ per $km^2$, or on average about 30 meters apart, and they occur more frequently in older terrain \cite{hiesinger2012old,minton2019equilibrium}. This would provide frequent landmarks that should be detectable at distances of roughly 10 to 20 meters from the rover.

Robotic lunar rovers will carry a sensor suite for relative localization and obstacle detection that includes wheel odometry, an IMU, either stereo cameras or a lidar, and either a sun sensor or a star camera for absolute heading measurement. Assuming the availability of a star camera, which is applicable for driving in sunlight and in shadow, gives 3-axis attitude knowledge to a small fraction of a degree. With this, the relative navigation sensor suite enables dead-reckoning with position error that typically can be $< 2\%$ of distance traveled. This provides a prior estimate of position that at all times strongly constrains which crater(s) from the landmark database are expected to be near the rover. Craters can then be detected near the rover with a combination of 3D point cloud data from stereo cameras or a lidar, image data from a camera, or reflectance image data from a lidar. This enables detecting craters with diameters roughly between 5 to 20 meters whose near rims are roughly less than 20 meters from the rover.

Overall, these methods should enable reliable absolute localization; given typical resolution characteristics of cameras, stereo vision, and lidar, we estimate that it should be possible to maintain a rover absolute position estimate with $3 \sigma$ error $< 5$ meters at all times. Furthermore, in terms of computational feasibility, crater-based localization is less expensive than obstacle detection and needs to be done much less frequently than obstacle detection, so any onboard computing system that can do obstacle detection would also be able to do crater-based localization.

\begin{table*}[!tbh]
    \centering
    \caption{\bf{Key Performance Parameters (KPP)}}
    \includegraphics[width=0.9\linewidth]{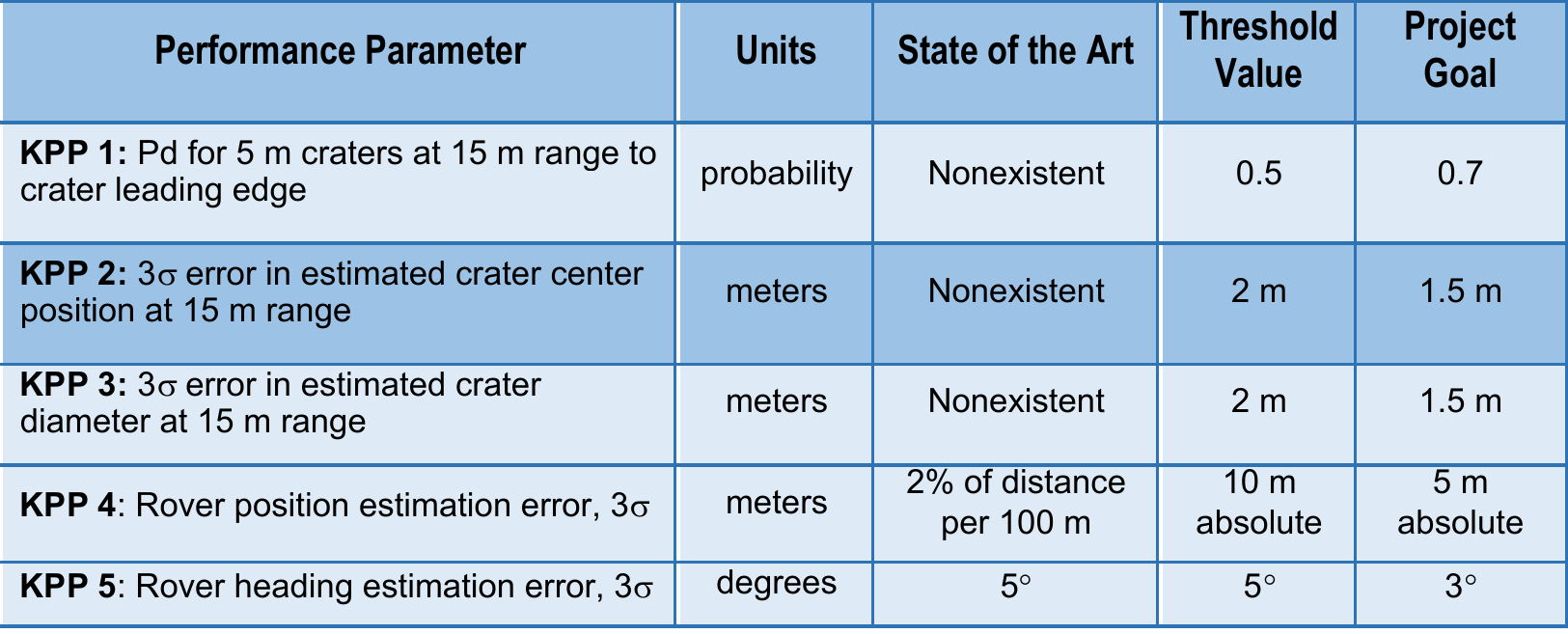}
    \label{tab:kpp}
\end{table*}

\subsection{Key Performance Parameters}
Thorough performance evaluation of the LunarNav framework was a function of many parameters, and depended fundamentally on the ability to detect and localize individual craters and to estimate their positions and diameters. This depends on (1) crater size and distance from the rover, (2) rover camera/lidar sensor parameters, including angular resolution, range resolution, field of view, and sensor height above the ground, (3) lighting conditions, and (4) other characteristics of terrain geometry, like slope. We distilled key performance parameters (KPPs) in terms of the crater distribution, as defined in Table \ref{tab:kpp} and anticipated that craters with diameters $>$ 5 meters will be readily detectable from distances of at least 15 meters, in many but not necessarily all lighting conditions. For example, detection probability ($P_d$) of 0.5 for 5 m craters at 15 m range should be a conservative estimate. With a notional stereo camera system with angular resolution of 1 milliradian/pixel and binocular camera baseline of 30 cm, 3 $\sigma$ errors in estimating the position and diameter of such craters should be $<$ 2 m each.

Updates to rover position and heading will be obtained every time an onboard crater map is registered to the orbital map. The precision of these rover position and heading estimates will be a function of the precision of crater positions and diameters in the onboard map, as well as detection and false alarm probabilities ($P_d$ and $P_f$) for onboard crater recognition. Quality of the onboard map will also depend on accuracy and precision of heading estimation and dead reckoning between crater detections. Past experience suggests that dead reckoned position error can be better than $2\%$ of distance traveled with visual odometry (2 m per 100 m) \cite{rankin2021mars}. Rover heading error should be bounded by about $5\degree$ by a sun sensor, $<1\degree$ per 100 m using visual odometry, or $<0.01 \degree$ at all times using a star camera. A star camera is the preferred heading estimation solution for performance, but a sun sensor may offer a lower cost solution with adequate performance for predominantly sunlit scenarios. Combining this with errors in crater center positions relative to the rover and crater position errors of $<$ 2 m in the orbital map should enable rover position and heading estimation error to be conservatively bounded by about 10 m and 5$\degree$ at all times. Erroneous crater detections (false alarms) will be filtered out through a combination of several techniques applied at different stages of the estimation pipeline. Since the effect of false detections is ultimately captured in rover position estimation error, a separate key performance parameter is not specified for false alarms. Performance modeling and evaluation throughout the course of the project characterized how performance varies as a function of these sensor parameters (See Section \ref{sec:exp}).


\section{Real and Simulated Datasets}
\vspace{2mm}
\begin{figure*}[!tbh]
    \centering
    \includegraphics[width=\linewidth]{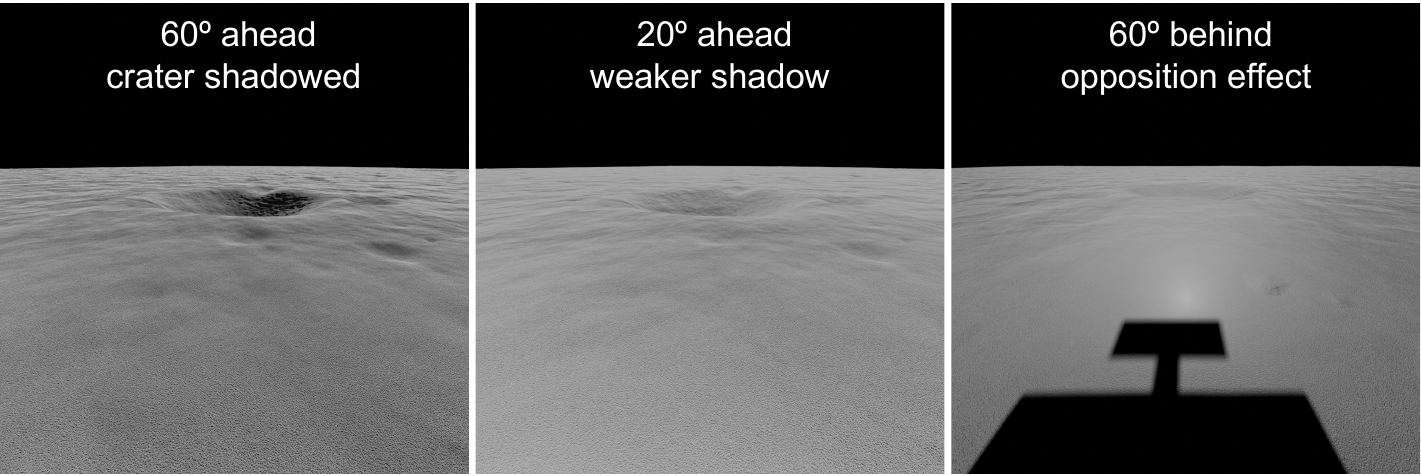}
    \caption{\textbf{Simulated images with different sun zenith angles to illustrate effect on image contrast.}}
    \label{fig:image_sim}
\end{figure*}

We used both real and simulated sensor data to develop and evaluate the performance of lunar rover navigation with crater landmarks. Simulated data allowed testing with larger data sets and over wider ranges of illumination conditions than are practical with real data. For generating synthetic lunar surface images, three different simulation tools were initially considered: (1) RSIM - a simulator implemented in the Gazebo environment at NASA Ames \cite{allan2019planetary}, (2) a lunar surface simulation environment under development in JPL’s DARTS lab to support the STMD/GCD “CADRE” task and other applications \cite{aiazzi2022iris}, and (3) an open-source Blender based computer graphics rendering engine \cite{matthies2022lunar}. RSIM had very low sun elevation angles rendered into terrain maps for use in simulating polar missions, and required changes to generalize the functionality to arbitrary sun elevation angles was beyond the scope of LunarNav timeline requirements. Similarly, the DARTS-based simulator was not scheduled for completion in the timeframe needed for LunarNav. Therefore, Blender was used to create synthetic images. Synthetic lidar images were generated similarly, by rendering 3D point clouds instead of images; sun angle was considered irrelevant for lidar, so that variable was ignored and we were able to use the higher fidelity RSIM. It is to be noted that both the Gazebo-based and DARTS-based simulators have more general capability, including simulating rover behavior, so this trade should be re-evaluated for end-to-end simulation needs in future years.

\subsection{Blender-based Lunar Scene Simulator}
The Blender-based image scene simulation takes a lunar digital elevation map (DEM), applies a custom texture map to it, and creates a world model with a simulated sun as a light source; within this model, a stereo camera pair is added and the simulation generates locations for the cameras and the sun. For each camera and sun location, Blender’s render engine produces a synthetic image. This image is rendered using a custom implementation of the Hapke radiometric model \cite{hapke1963theoretical,hapke1981bidirectional,hapke1993opposition} incorporated into the Blender’s path tracing algorithm to simulate accurate lighting across the surface. The simulation has multiple parameters that can be controlled, including sun position, camera extrinsic and intrinsic parameters, image size, and a few others. Images were were generated for craters with diameters ranging from 5 to 20 meters, with four different approach angles for each crater. For each crater and approach angle, the stereo camera was positioned at distances between 5 and 20 meters from the crater near rim, at 1 meter spacing; for each location, an image was rendered with a varied set of sun angles from 0$^{\circ}$ to 80$^{\circ}$ from nadir. The final dataset consisted of 1,792 stereo pairs.

\begin{figure}[!tbh]
    \centering
    \includegraphics[width=\linewidth]{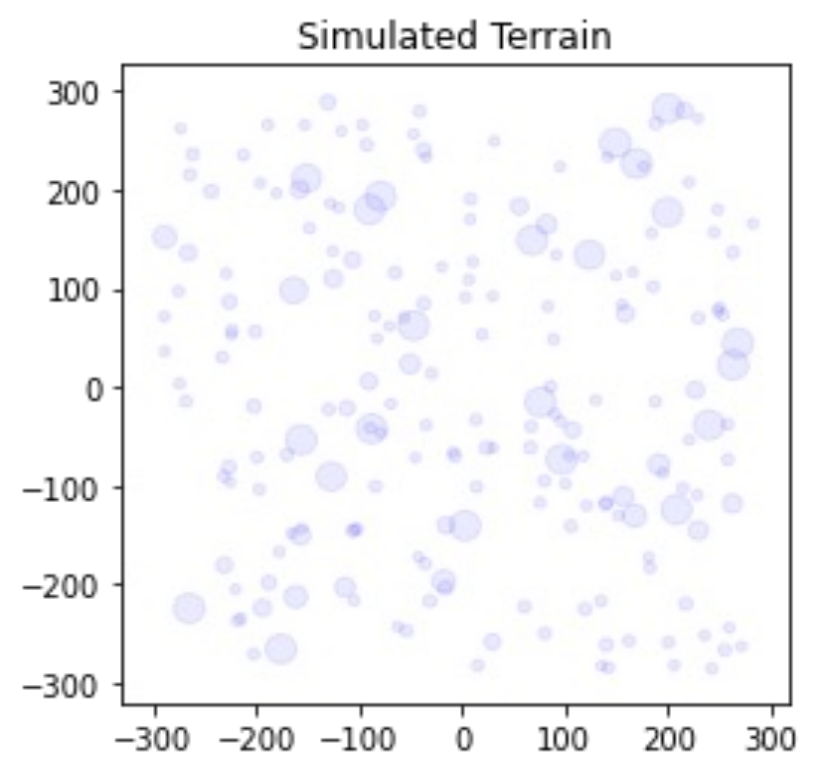}
    \caption{\textbf{Illustration of the simulated environment used for development and testing of localization algorithms. Shown here is an example for a 600m x 600m region with known orbital craters}}
    \label{fig:mc_sim_env}
\end{figure}

\subsection{2.5D Simulation Environment for Long-range Traverse}
To facilitate development of algorithms for crater matching based localization, we constructed a 2.5D simulation environment that allows for independent unit testing of the different localization algorithms. This allowed us to quickly iterate through different versions of the localization algorithms and validate their effectiveness with offline Monte Carlo simulations. In the simulated environment, the localization algorithm receives two sets of craters as input: the known craters detected from orbit and the ones observed by the rover from the ground. The orbital craters are parameterized by their size and location (x, y, diameter). Their craters sizes are drawn from a truncated power-law distribution ($\alpha$ = 1, diameter $<$ 20), whereas their locations are drawn from a uniform distribution. The craters observed on the surface by the rover are generated by perturbing the known orbital craters according to a similar distribution of noise from crater detection, with a zero-centered gaussian ($\sigma$ = 3m) error in the crater location and zero-centered gaussian ($\sigma$ = 1m) error in the crater size. Further, we also added the capability to mask a percentage of either orbital or ground craters to simulate false positives and false negatives in crater detection. Figure \ref{fig:mc_sim_env} shows an illustration of the simulated environment that was used. Assuming the motion model with 2$\%$ noise, the rover traverses within this simulated environment and observes craters within a field of view range ($<$ 40m). The localization algorithms then match the observed ground craters to the orbital craters for localization. Then, the estimated route is compared against the ground truth route for evaluation. The simulated environment can be extended to long-range traversals and can be used to validate the effectiveness in overcoming the baseline $2\%$ of motion model noise.

\begin{figure}[!t]
    \centering
    \includegraphics[width=\linewidth]{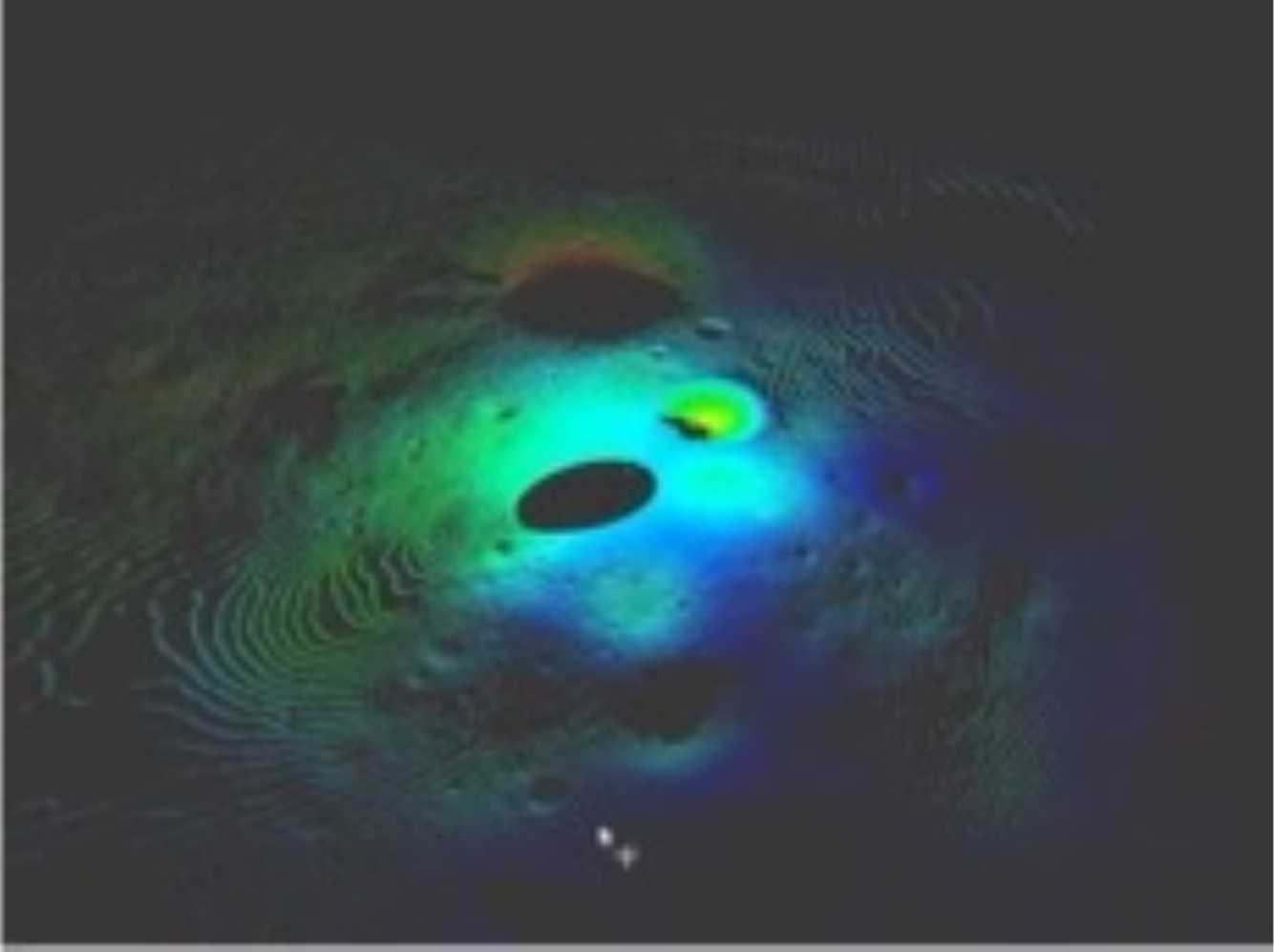}
    \caption{\textbf{Illustration of an instance from the LiDAR simulation from RSIM.}}
    \label{fig:lidar_sim}
\end{figure}

\begin{figure*}[!t]
    \centering
    \includegraphics[width=\linewidth]{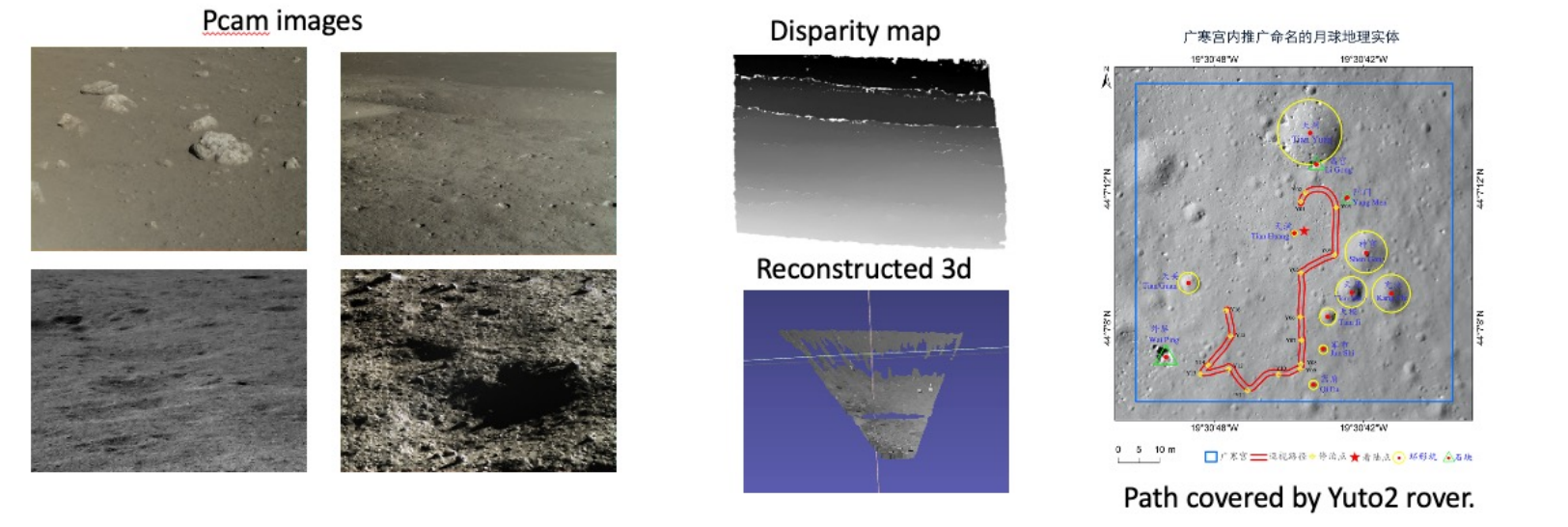}
    \caption{\textbf{Example Chang’e 3 Yutu rover images (left), an example stereo disparity map and 3D rendering (middle), and orbital image showing the rover traverse (right). Stereo disparity is inversely proportional to range; in the disparity map, bright pixels are close and dark are far.}}
    \label{fig:change-stereo}
\end{figure*}

\begin{figure*}[!t]
    \centering
    \includegraphics[width=0.95\linewidth]{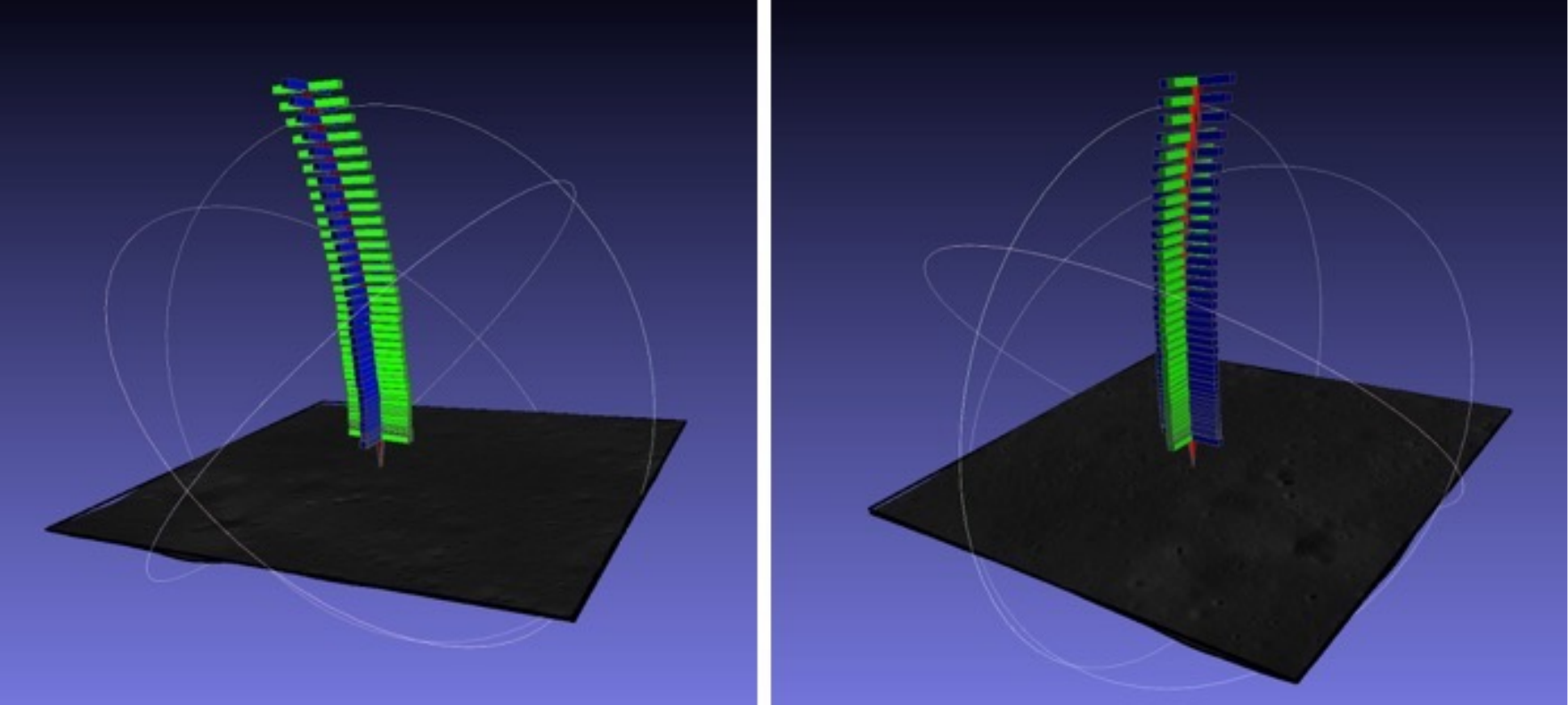}
    \caption{\textbf{The Chang’e 4 EDL Flight trajectory was recovered between 1000m and 100m above ground, using TRN algorithm on the LCAM images.}}
    \label{fig:change-lcam}
\end{figure*}

\begin{figure*}[!t]
    \centering
    \includegraphics[width=0.95\linewidth]{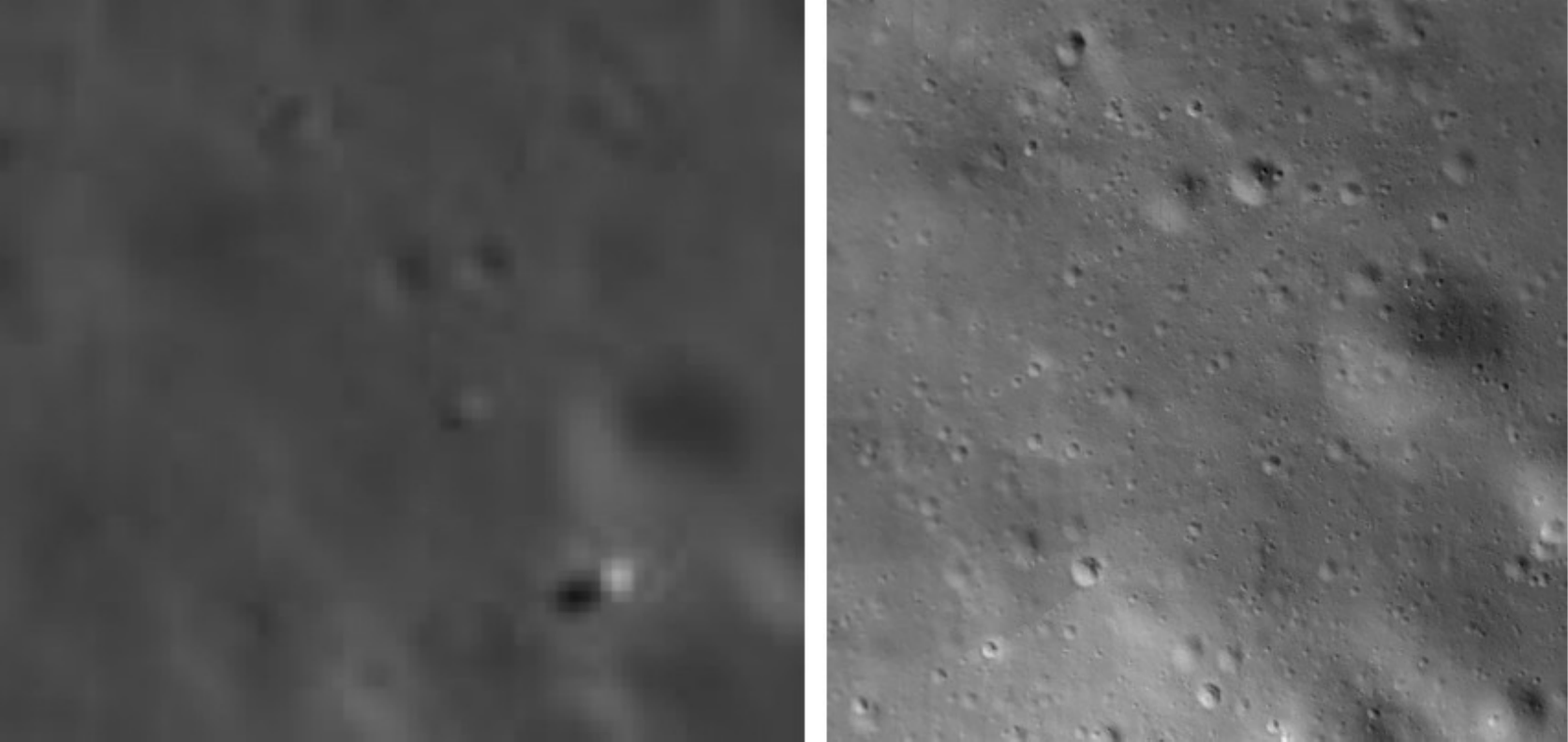}
    \caption{\textbf{Comparison of (Left) Original resolution - 1.4 m/pixel vs (Right) Improved resolution – 0.2 m/pixel for the orbital maps for the Chang’e 4 landing site.}}
    \label{fig:change-improved}
\end{figure*}

\begin{figure*}[!t]
    \centering
    \includegraphics[width=\linewidth]{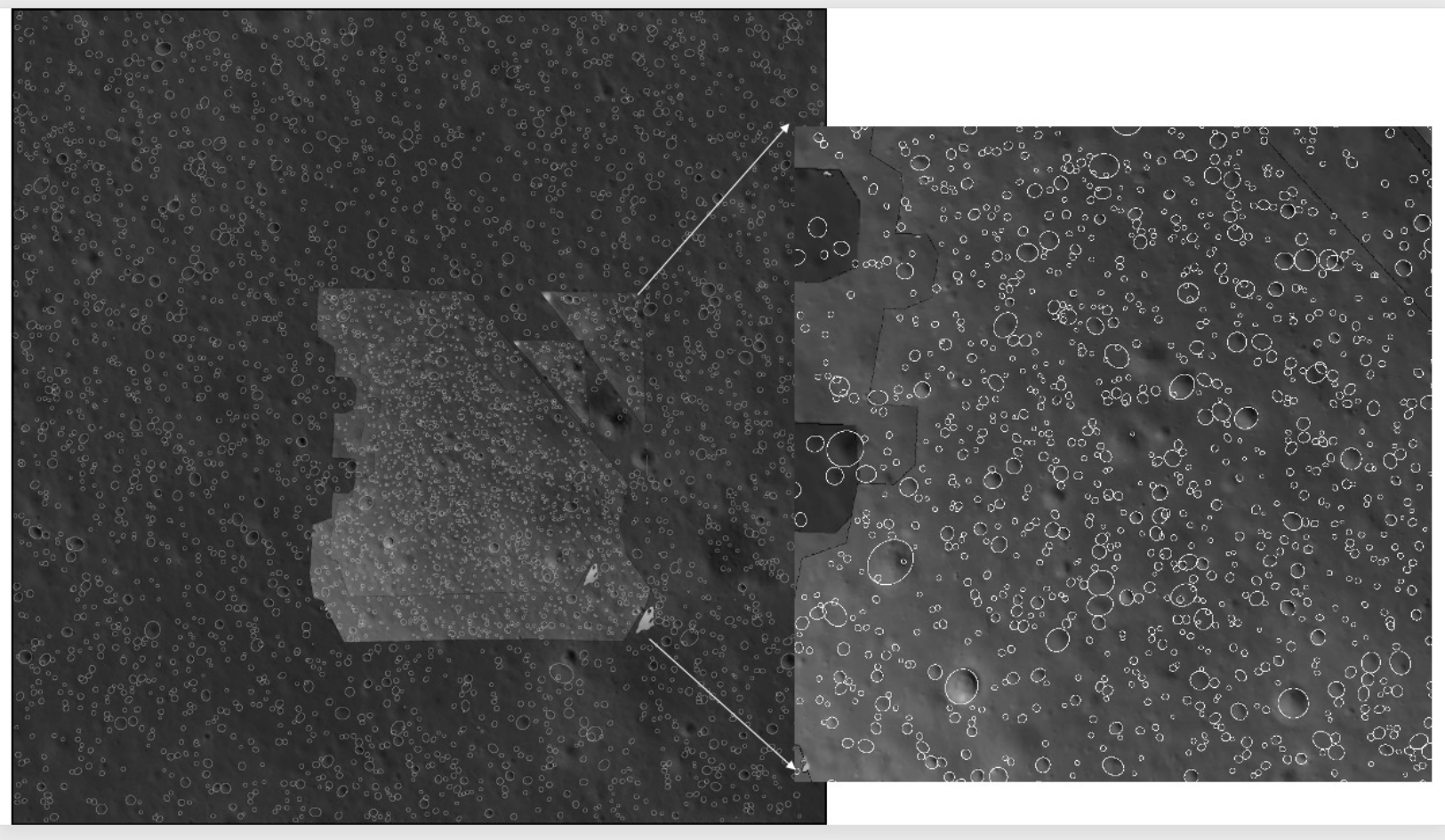}
    \caption{\textbf{Crater landmark map generation combining LRO-NAC images with higher resolution Chang’e LCAM descent imagery. Large image is from LRO-NAC; smaller overlay and the expanded view on the right is from LCAM.}}
    \label{fig:change-crater}
\end{figure*}

\subsection{Long-range LiDAR simulation using RSIM}
To further simulate long-range traverses from real Lunar terrain, we added a LiDAR simulation capability in RSIM \cite{allan2019planetary}, a ROS2 and Gazebo based high-fidelity simulation environment developed at NASA Ames to support the development of the VIPER Rover \cite{colaprete2019overview}. RSIM models the VIPER lunar rover structure, with the full CAD including rover kinematics, dynamics, and 3D model. RSIM also contains a lunar terrain model of the Nobile landing site, with 8k resolution. The terrain includes rocks and crater distributions, shadowing, and sun lighting. The LiDAR sensor (also referred to as the Velodyne Simulator), is forked from the Toyota Research Institute Velodyne Simulator. This package contains a URDF description and Gazebo plugins to simulate the Velodyne laser scanners (our model is the HDL-32E). This Gazebo plugin is based on the gazebo plugins ROS block laser, which publishes PointCloud2 messages with the same structure (x, y, z, intensity, ring) and simulates Gaussian noise. To capture the point clouds from the LiDAR sensor, we created a package that provides a ROS2 node that can subscribe to the Velodyne lidar sensor mounted on the VIPER rover and saves a snapshot to the desk. In addition to capturing the point cloud, this node also outputs the position and orientation of the rover with respect to the world origin. It also contains scripts to visualize the point cloud and export it to other formats. The final datasets used were two discrete drives of the VIPER rover in RSIM with LiDAR on the Nobile landing site; both drives started the rover at the bottom-left corner of the map (320m x 320m) and traversed diagonally along the hypotenuse to the upper-right corner, took approximately 30-32 steps of 10m each, with a change in orientation chosen randomly between (-5 to 5 degrees) at each step. Due to the orientation change, the rover could not complete all the steps as it would fall off the map since it was not driving in a purely straight path with step-length: 10m, total steps: 35 and orientation change: 5 degrees at every waypoint. Figure \ref{fig:lidar_sim} shows an illustration of an instance the LiDAR simulation from RSIM. 

\subsection{Real Lunar data from Chang'e missions}
For real sensor data, the original plan for to acquire a large- scale dataset of LIDAR and stereo image data from an analog site at the Cinder Lake crater field near Flagstaff, Arizona. This had to be postponed due to covid travel restrictions and forest fires; in lieu of this data, we decided to use real lunar stereo images that were available from the “panoramic” cameras (PCAM) on the Chang’e 3 \cite{li2015chang} and Chang’e 4 \cite{li2021overview} lunar rover missions. Since this is actual lunar data, in important ways it is better than the analogue stereo images that were originally planned. The Chang’e missions’ rover data has been publicly released \footnote{https://moon.bao.ac.cn/}. The cameras for both missions are identical; their FOV is 19.7$^{\circ}$x14.5$^{\circ}$, resolution are 2352 x 1728 pixels (color) and 1176 x 864 pixels (monochrome) and stereo baseline length is 27 cm. A total of 168 pairs of Chang’e 3 and 1174 pairs of Chang’e 4 PCAM images were processed. A stereo camera self-calibration applied to this data yielded good camera models for both data sets. Stereo depth maps were computed from these images with good results, as shown in Figures \ref{fig:change-stereo}. 

Ground truth labeling of craters in this imagery and registration of this imagery against orbital imagery was done to prepare a dataset that was used for performance evaluations. Towards this end, we studied the usefulness of the Chang’e lander camera (LCAM) to obtain a high resolution and high precision crater database. The LCAM image sequences (5000 images) were downloaded, and the Chang’e EDL flight trajectory was recovered between 1000m and 100m above ground, using terrain relative navigation (TRN) and structure from motion algorithms. Figure \ref{fig:change-lcam} shows visualization of the recovered LCAM trajectory. Then, the LCAM images were ortho-rectified to a coarser resolution LRO-NAC image map (1 m/pixel) to obtain an ortho-image with higher resolution of up to 20 cm/pixel around the lander. This process was used to improve the resolution for a 100m x 100m region around the Chang’e 4 lander as shown in Figure \ref{fig:change-improved}. The craters were detected from this high-resolution map and their geographic locations, diameters, depths were extracted into a crater database, as shown in Figure \ref{fig:change-crater}. This database was used for rover localization algorithm development
%



\label{sec:data}

\section{Technical Approach}
\vspace{2mm}
LunarNav's technical approach consists of three main elements: 1) crater detection, 2) crater matching, and 3) state estimation. In previous work, we
developed crater detection algorithms for three different sensing modalities \cite{matthies2022lunar}. This paper builds on that work, and focuses on the
crater-based localization (i.e. crater matching and state estimation) aspect of the problem. We will start with a brief review of how we do crater detection, and then introduce two crater-matching based localization algorithms that we developed: parametric and non-parametric methods. 

Each approach has both advantages and disadvantages; non-parametric approaches like the particle filter make no assumptions about the probability distribution for rover location and are robust in a wide variety of scenarios. However, the run-time may be slow due to the large number of particles it uses to represent the possible states of the rover. In contrast, parametric approaches are faster and more computationally efficient because rover position uncertainty is represented by closed-form mathematical expressions. Because the parametric approach makes certain assumptions about the distribution of errors, it may be less robust compared to non-parametric approaches. 

\subsection{Review of Crater Detection}
Planetary rovers to date have all used stereo cameras for terrain imaging and 3D perception \cite{goldberg2002stereo}. Future rovers (to Moon and Mars) might have LIDAR, either by itself or in addition to one or more cameras. Thus, any combination of these sensors could be used for crater detection. To cover this set of options and to create a foundation for identifying the best approach in the future, crater detection algorithms were developed for three classes of technical approach: (1) Using 3D point clouds from LIDAR, (2) Using 3D point clouds from stereo vision, and (3) Using deep-learning based pattern recognition with monocular images. 

The original plan was to treat crater detection as a process done independently from any knowledge of the crater landmark map and rover position; however, work last year showed that more reliable crater detection results was achieved by assuming approximate prior knowledge of rover position, which is realistic in practice, and using that to allow the crater detection process to invoke 3D models of craters expected to be around the rover based on this prior knowledge. Such approximate prior knowledge was used for the LIDAR- based approach developed, but has not yet been carried over to the other approaches. Geometric analysis methods was appointed to the point clouds from LIDAR and stereo vision; machine learning with neural nets was used for monocular images.

Results of quantitative performance evaluation with geometric analysis applied to simulated 3D point clouds from LIDAR show high reliability for detecting craters with a leading edge within about 15 m from the rover. The results also suggested that rover localization with an error less than 5 m is highly probable. Somewhat simpler geometric analysis methods were applied to simulated 3D point clouds from stereo vision, which are noisier than LIDAR-based point clouds. Stereo-based detection degraded at shorter range than for LIDAR and obtained significantly higher crater position estimation error; nevertheless, rover localization with error in the 5 m range still appears to be possible. Monocular appearance-based detection was done with a CNN-based machine learning algorithm; this produced detection results in image space, but did not produce 3D crater position and size estimates. Detection performance exceeded the other two methods, making this a very promising approach for future crater-based localization systems. See \cite{matthies2022lunar} for details on the algorithms and performance characterization of crater detection.

\begin{figure}
    \centering
    \includegraphics[width=\linewidth]{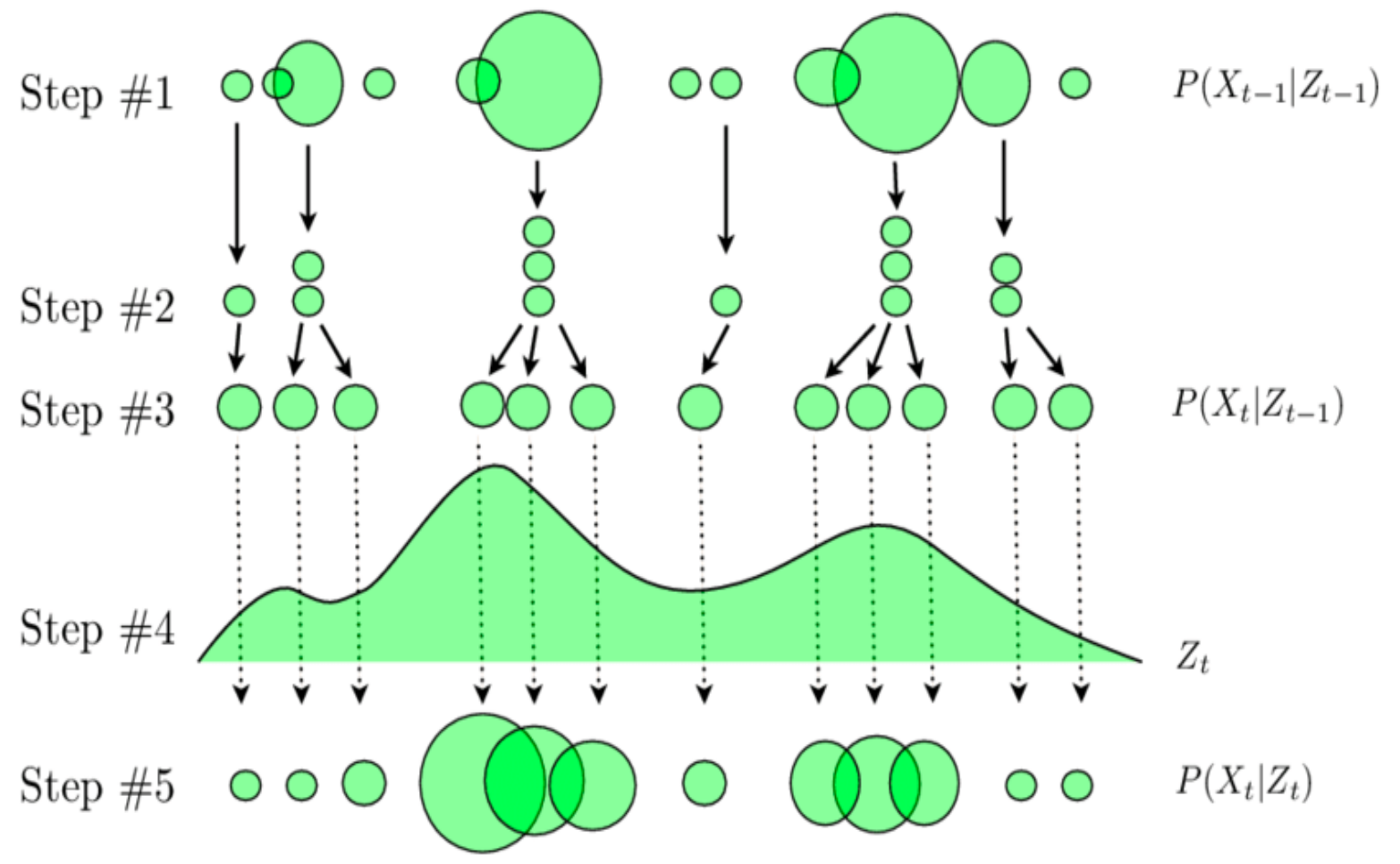}
    \vspace{2mm}
    \caption{\textbf{Overview of the Particle Filter framework. Steps associated with a particle filter approach. (1) The particles associated with the a posteriori function t-1 are (2) sampled; and (3) propagated leading to the a priori function P($X_t|Z_{(t-1)}$). Then, upon (4) observation in t, we have the (5) a posteriori function P($X_t|Z_t$)in t.}}
    \label{fig:partcile_filter}
\end{figure}

\begin{figure*}[!t]
    \centering
    \includegraphics[width=0.7\linewidth]{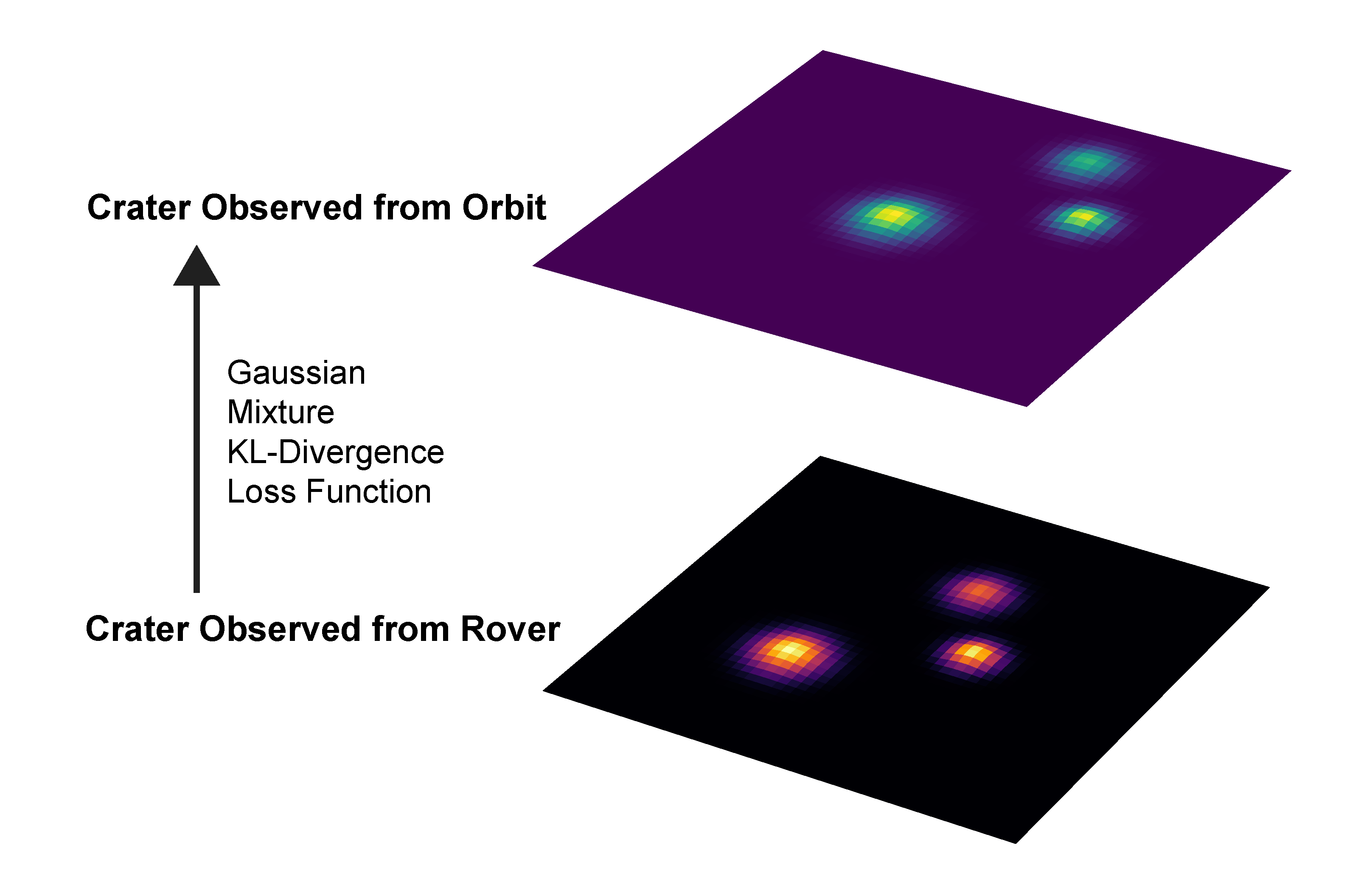}
    \caption{\textbf{Formulating rover and orbital observations as gaussian mixture models. Using optimization to find the best match between two closed-form terrain models for localization.}}
    \vspace{5mm}
    \label{fig:gmm}
\end{figure*}

\begin{figure*}[!t]
    \centering
    \includegraphics[width=\linewidth]{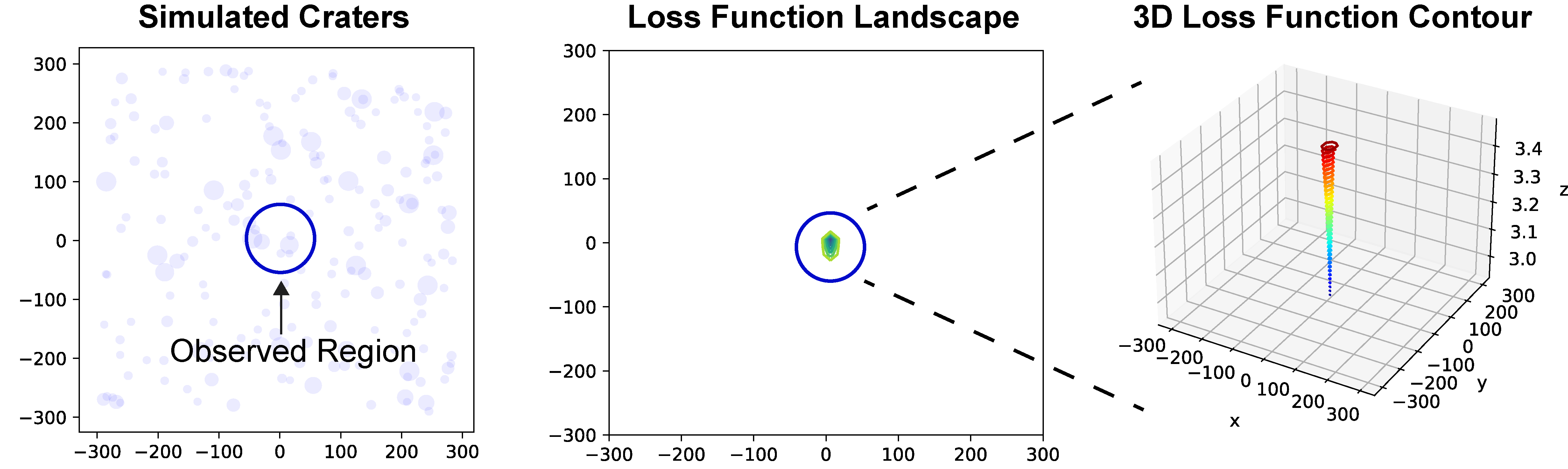}
    \caption{\textbf{Parametric loss function landscape of the gaussian mixture models.}}
    \vspace{2mm}
    \label{fig:loss_landscape}
\end{figure*}

\subsection{Crater-based Localization using Particle Filters}
For non-parametric methods, we implemented a crater-based localization algorithm based on Particle Filters \cite{thrun2002particle,fox2001particle}, where each particle represents a hypothesis for the rover location, and the strength of the hypothesis is represented by a measurement probability. Figure \ref{fig:partcile_filter} illustrates a generalized overview of the particle filter’s cycle in which we have, in the first layer, some particles represented by ellipsis of various sizes. The size denotes the weight of a particle. The second layer illustrates the result of the sampling process which can lead to repeated particles. Upon sampling, the weights of the particles lose their meaning and new measurements are necessary as the third layer shows. In this layer, the particles’ weights are adjusted according to the used motion model and a noise model applied to the sorted particles. Upon adjustments, we have an estimation for the next frame in the sequence. The fourth layer shows the function representing the updated particles. In this function, the height denotes the weight of the measurement at a given point. The fifth layer shows the a posteriori probability function as the result of the measurement step. In this step, the particles are properly weighted and prepared for the next iteration of the localization cycle.

In our implementation, a large number of particles can be used to approximate the distribution of the rover location as the rover moves and detects the craters. To estimate the location, the particles are moved according to the motion model, then the measurement probabilities are updated by comparing the ground craters with the orbital craters. Lastly, the particles are re-sampled according to their measurement probabilities to represent the new location distribution after motion. 

There are many possible formulations on how to update the measurement probability given certain observations; we reasoned that the geometric relationship between a set of crater observations can be used to improve the accuracy of localization, instead of modeling each crater as an independent observation. Therefore, we constructed a spatial formulation where the measurement probability of each particle is represented by the average area Intersection Over Union (IoU) distance \cite{rezatofighi2019generalized} between the orbital and ground craters:

\begin{figure}[h]
    \centering
    \includegraphics[width=\linewidth]{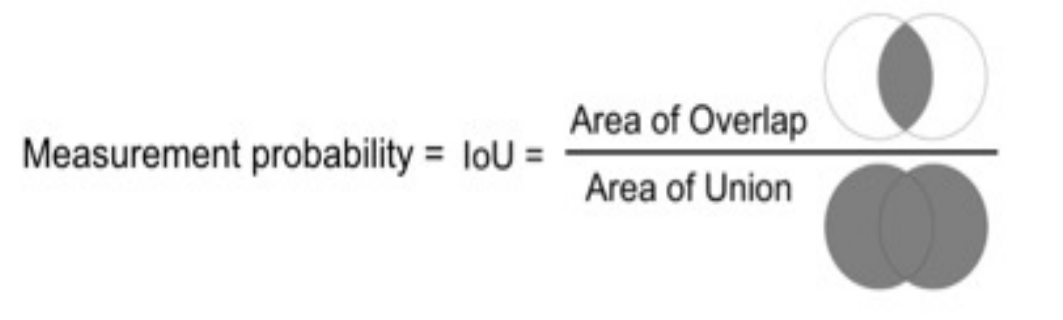}
\end{figure}

To account for missed detections, we adjust for the probability of new crater detection by using prior probability of detection from previous data.

\begin{figure*}[!t]
    \centering
    \begin{tabular}{cc}
    \includegraphics[width=0.5\linewidth]{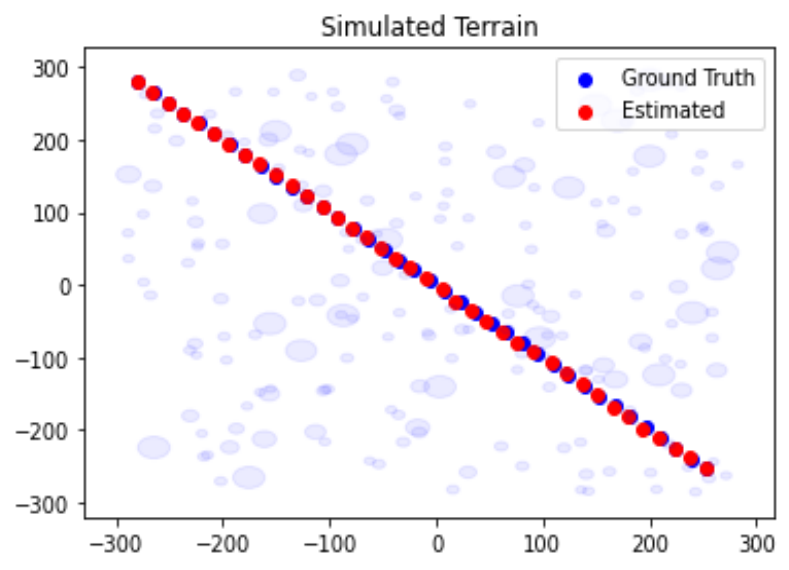} &
    \includegraphics[width=0.5\linewidth]{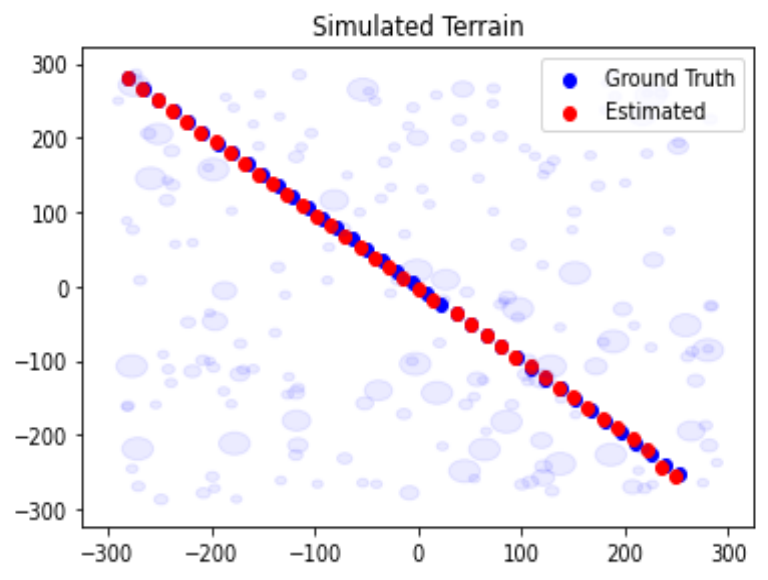}
    \end{tabular}
    \caption{\textbf{(Left) Example of particle filter performance on a simulated scenario, and (Right) Example of parametric model performance on a simulated scenario.}}
    \label{fig:lidar_res1}
\end{figure*}

\begin{figure*}[!t]
    \centering
    \begin{tabular}{ccc}
    \includegraphics[width=0.31\linewidth]{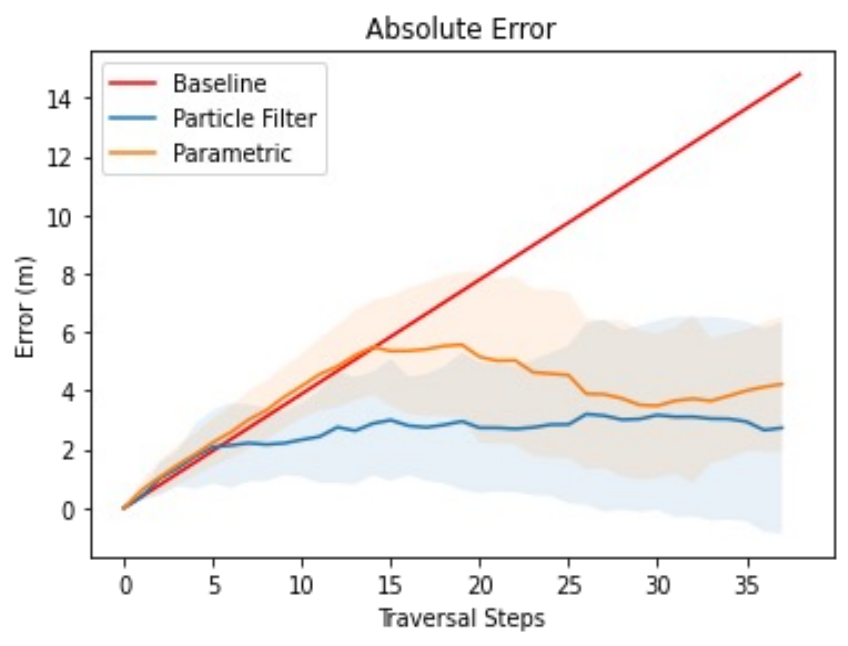} &
    \includegraphics[width=0.31\linewidth]{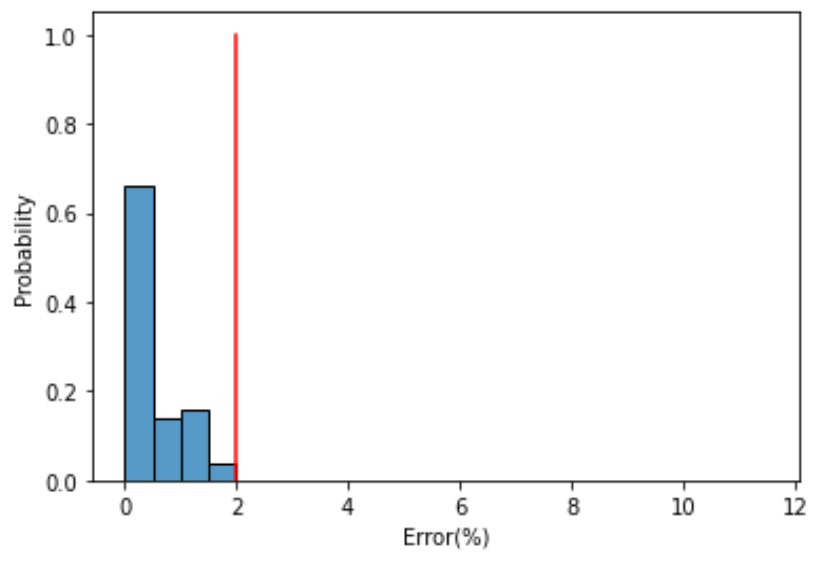} &
    \includegraphics[width=0.31\linewidth]{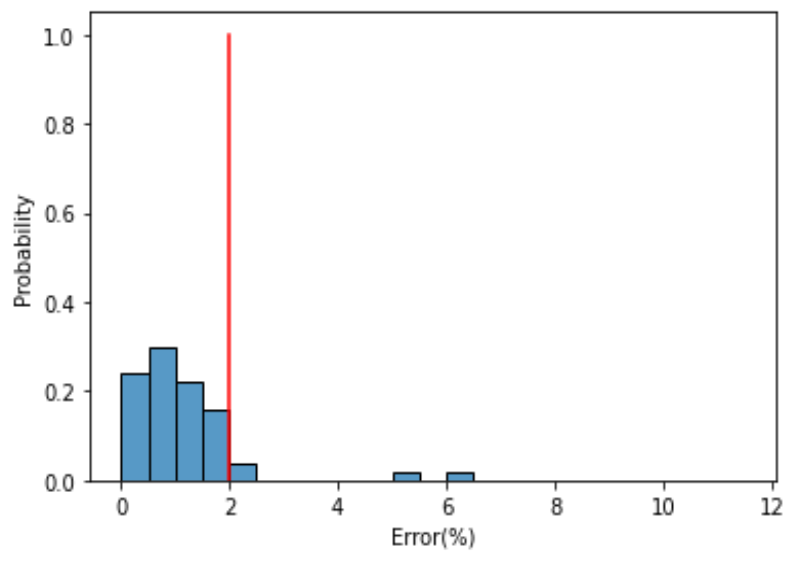}
    \end{tabular}
    \caption{\textbf{(Left) Performance of particle filter and parametric model over different (n=50) simulated environments. The red line indicates the 2$\%$ noise in motion model baseline, and the distribution of the relative error at the end of the traversal for particle filter (Middle) and parametric model (Right) with the 2$\%$ motion model baseline in red.}}
    \label{fig:lidar_res2}
\end{figure*}

\subsection{Crater-based based Localization using Parametric Matching}
For parametric method-based localization, we assumed that each crater is represented by a gaussian distribution, with the location of the crater as the mean and the half of the radius as the standard deviation. For orbital craters, the set of craters can then be expressed using a gaussian mixture; and this can also be independently formulated for ground craters. Parametric matching based localization is formulated as the problem of finding the translation between the two gaussian mixtures that best matches these two distributions as shown in \ref{fig:gmm}. The KL-divergence \cite{goldberger2003efficient} measures the distance between two distributions and can be used as the loss function. We minimize the loss function with gradient-descent based methods to find the best translation, which represents the location (as illustrated in Figure \ref{fig:loss_landscape}). Further, the Hessian around the optimal solution can be used to estimate the standard error, since KL-divergence is equivalent to the negative log-likelihood.

\section{Performance Evaluation}
\begin{figure*}
    \centering
    \begin{tabular}{cc}
    \includegraphics[width=0.5\linewidth]{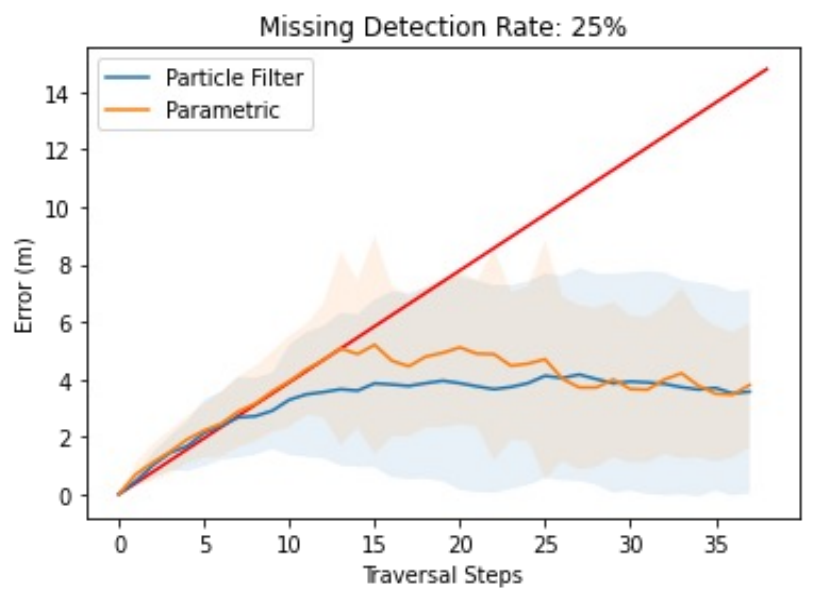} &
    \includegraphics[width=0.5\linewidth]{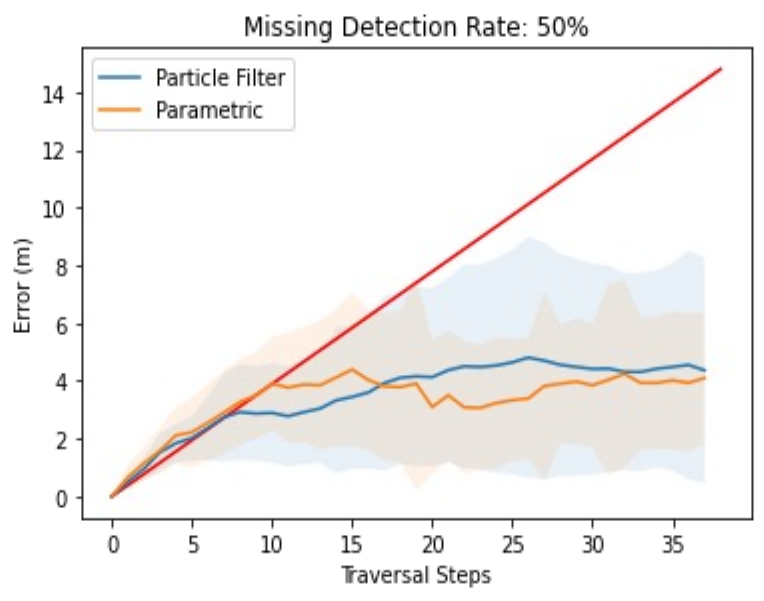}
    \end{tabular}
    \caption{\textbf{Performance of localization algorithms with 25$\%$ (Left) and 50$\%$ (Right) of the orbital craters masked, along with the 2$\%$ motion model baseline in red.}}
    \label{fig:lidar_res3}
\end{figure*}

\subsection{LIDAR-based Localization on Simulation Environment:}
We evaluated the particle filter and parametric localization algorithms on a 400m x 400m simulated environment with 100 craters. An example of the traversal and localization results are shown below in Figures \ref{fig:lidar_res1}-\ref{fig:lidar_res3}. Over a large number of simulated scenarios (n=50), both localization methods were able to perform better than the relative localization baseline of 2\% noise motion model, with the particle filter performing slightly better. Further, the particle filter converged at around 2m of error over long ranges, suggesting that crater landmarks are valid features for accurate localization. 

\begin{figure*}
    \centering
    \begin{tabular}{cc}
    \includegraphics[width=0.65\linewidth]{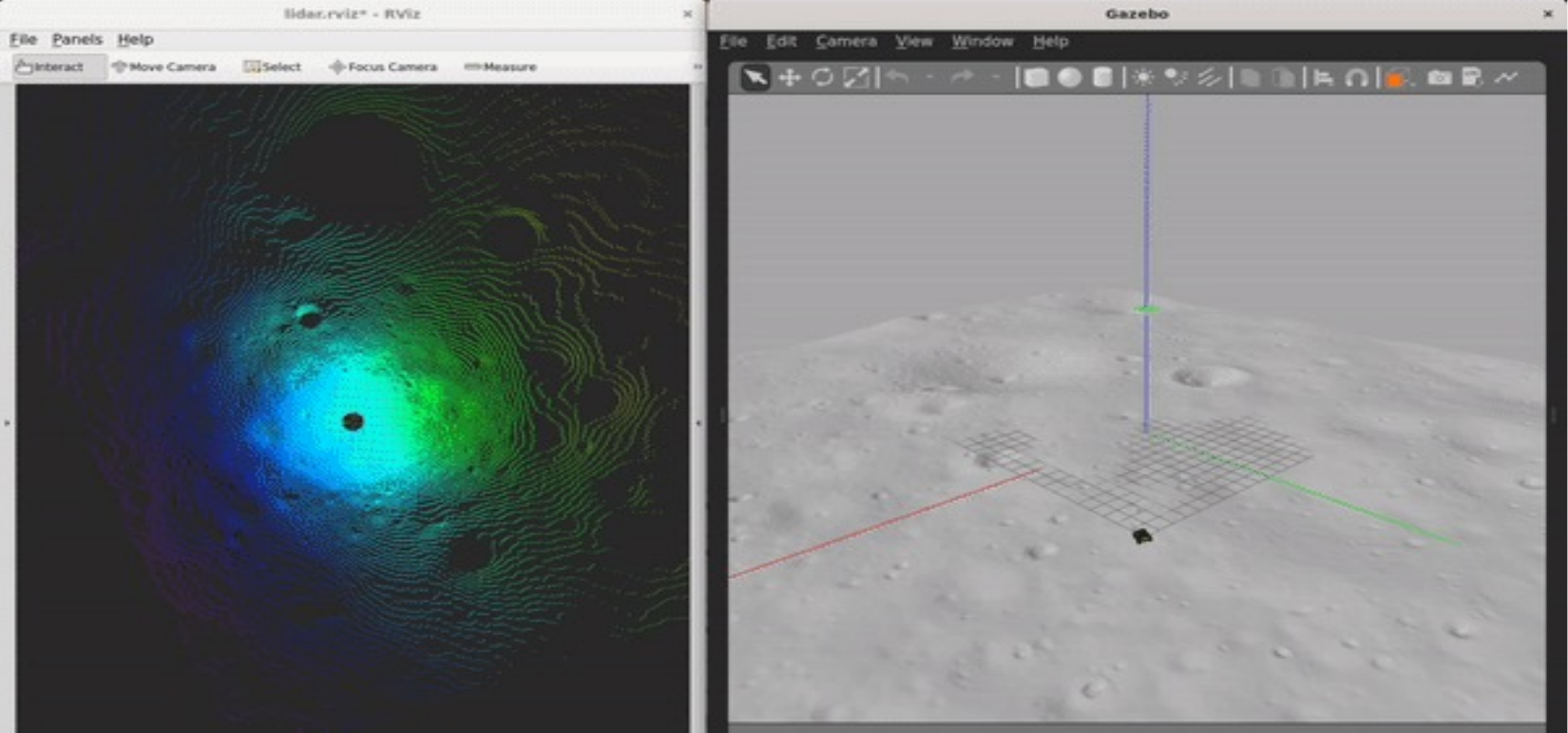} &
    \includegraphics[width=0.3\linewidth]{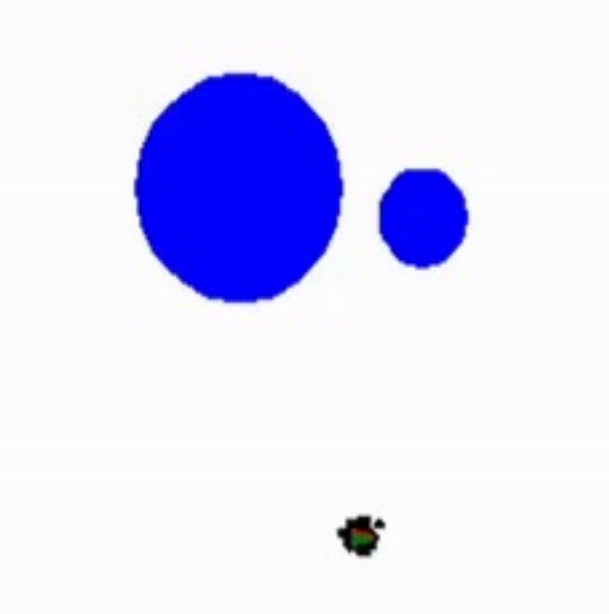}
    \end{tabular}
    \caption{\textbf{Illustration of the rover traverse for the Apollo 2 landing site showing the (left) simulated LiDAR point cloud and (middle) DEM overlayed with the rover traverse. Note: Point cloud is color coded using elevation data. (Right) An instance of particle filter illustration. Blue circles indicate craters, green arrow indicates estimated position, red arrow indicates GT position, and black dots represent the distribution of particles.}}
    \label{fig:lidar_apollo}
\end{figure*}

\subsection{LIDAR-based Localization from Apollo 2 Landing Site:}
We performed preliminarily tests of the particle filter and parametric localization on craters detected from simulated LIDAR point clouds. Here, we used the elevation map from the Apollo 2 landing site to simulate a traversal of 20 steps (with 1m/step) for crater detection. In this scenario, two craters (with diameter 6.20m and 14.76m) were manually labeled as orbital craters. The crater detection algorithm matches observed candidate craters on the ground with the two orbital craters, and the localization algorithm used the location of the observed craters for localization, as shown in Figure \ref{fig:lidar_apollo}.

\begin{figure*}
    \centering
    \begin{tabular}{cc}
    \includegraphics[width=0.5\linewidth]{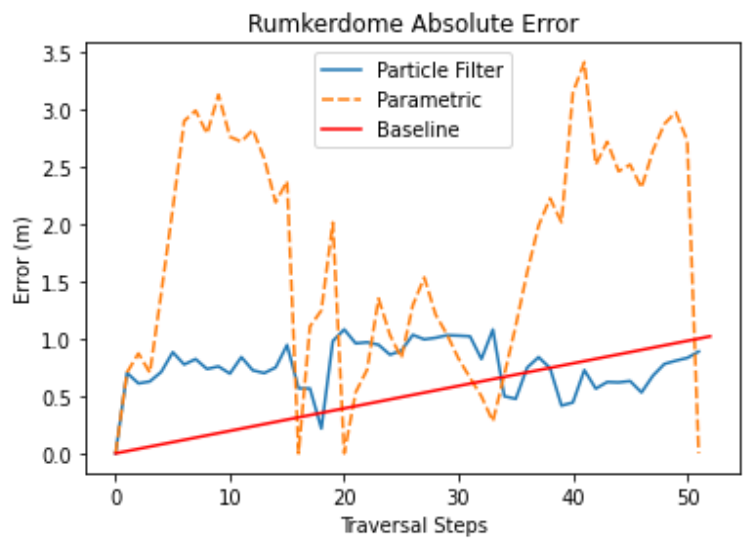} &
    \includegraphics[width=0.5\linewidth]{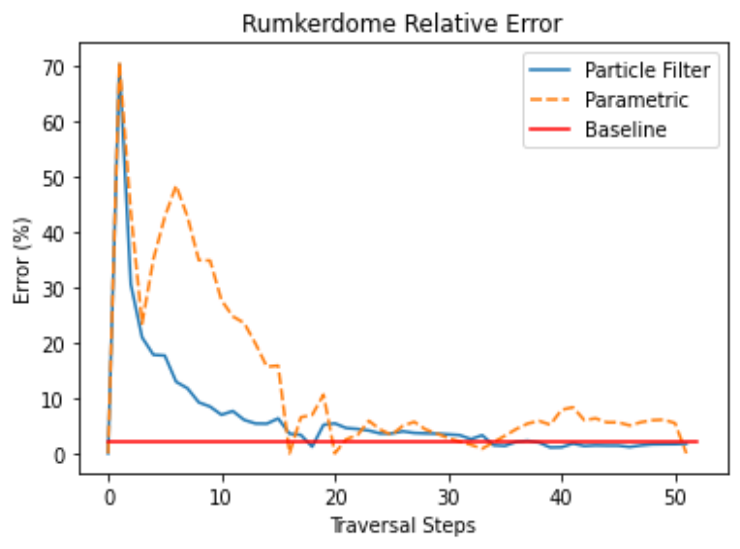}
    \end{tabular}
    \caption{\textbf{Rumker Dome site localization algorithm performance. The particle filter converged at around 0.75m of error, which is lower than the expected 2$\%$ (1m) motion model noise. The parametric method has an average error of 1.6m.}}
    \label{fig:lidar_rumker}
\end{figure*}

\subsection{LIDAR-based Localization from Rumker Dome Landing Site:}
Further, we evaluated the particle filter and parametric localization on LIDAR crater detected from a traverse through the Rumker Dome region. This is a more complex scenario compared to the Apollo 2 landing site, with 4 manually labeled craters (with diameter 21.94m, 5.60m, 6.98m, 8.58m) and 50 traversal steps (1m/step). According to the 2\% motion model baseline, our model is estimated to deviate 1m from the ground truth. Our results (Figure \ref{fig:lidar_rumker}) indicate an average distance of 0.75m throughout the traverse, indicating that crater landmarks are a reliable feature for more accurate localization. 

\begin{figure*}
    \centering
    \begin{tabular}{cc}
    \includegraphics[width=0.45\linewidth]{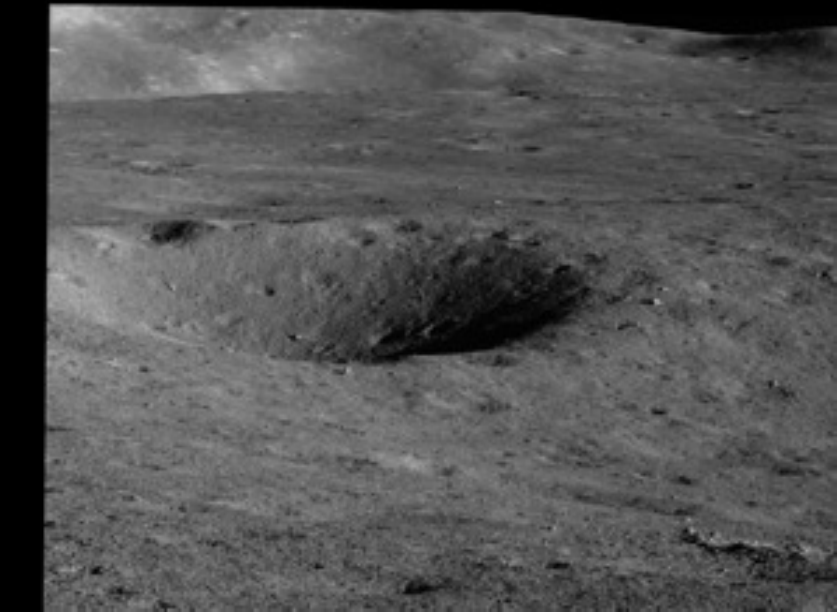} &
    \includegraphics[width=0.45\linewidth]{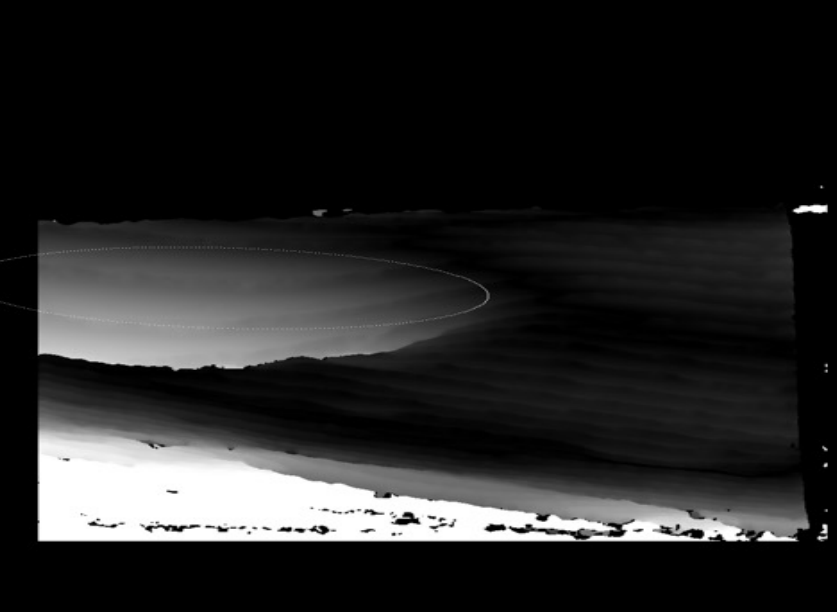}
    \end{tabular}
    \caption{\textbf{An example of a successful crater detection. (Left) original image; (b) disparity map overlayed with white ellipse showing crater detection}}
    \label{fig:stereo-change}
\end{figure*}

\begin{figure*}
    \centering
    \begin{tabular}{ccc}
    \includegraphics[width=0.31\linewidth]{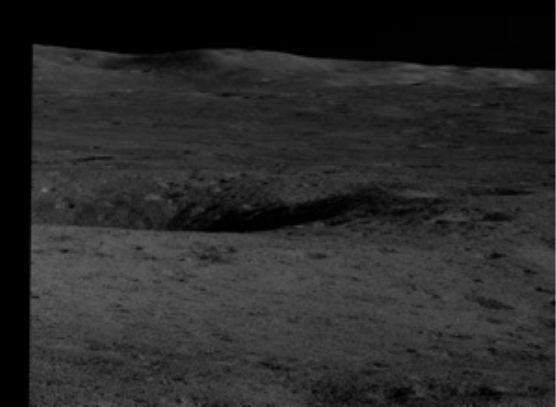} &
    \includegraphics[width=0.31\linewidth]{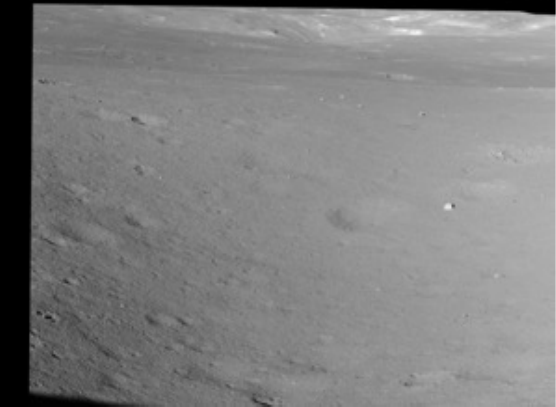} &
    \includegraphics[width=0.31\linewidth]{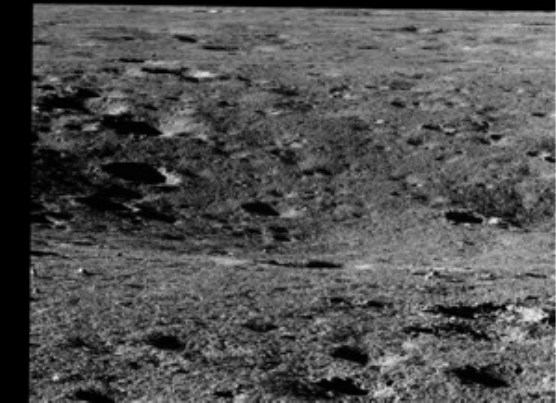}
    \end{tabular}
    \caption{\textbf{Example images where the stereo-based crater detection algorithm fails.}}
    \label{fig:stereo-fail}
\end{figure*}

\subsection{Stereo-based Localization on Chang'e 4 Landing Site}
We integrated the particle-filter-based localization algorithm with the stereo-based crater detection from FY21, and tested it with the Chang’e rover stereo data. The stereo crater detection algorithm succeeds in detecting craters on the real data when the assumptions of the algorithm \cite{matthies2022lunar} hold true; for example, as shown in Figure \ref{fig:stereo-change} which demonstrated a successful case of stereo-based crater detection on real Lunar data. 

However, the original crater detection makes two critical assumptions which regularly do not hold in the Chang’e 4 dataset. It first assumes that the entire crater is visible in the image. Due to the small field of view of the PCAM, this is often not the case in the Chang’E imagery. The second assumption is that the crater composes a relatively small section of the image, since otherwise the ground plane will not be detected correctly. This is also not the case with nearby craters, in part due once again to the small camera field of view. 

Figure \ref{fig:stereo-fail} shows three example images where crater detection currently fails. In the leftmost image, the algorithm successfully detects the as a crater candidate in the penultimate step of the algorithm. However, it is filtered out as not a crater because the entire crater is not visible in the image, violating an assumption behind the algorithm. In the middle image, the crater is not entirely within the image. This case fails in step three of the algorithm, earlier than the previous case, since the critical near rim edge is not visible. Furthermore, since the crater takes up almost the entirety of the image, the back wall of the crater is detected as the ground plane. In the rightmost image, the crater takes up a large section of the image, and the crater’s back wall is detected as the ground plane, leading the algorithm to fail. Additionally, the left and right edges of the crater are not visible so this would likely fail in a later step as well. Thus, the performance of the crater detection did not generalize to the current dataset of craters visible in Chang’e imagery. As a result, we were unable to benchmark the performance of the particle filter on stereo modality.

\label{sec:exp}

\section{Discussion}
\vspace{2mm}
The LunarNav project made the following accomplishments and contributions to crater-based localization:
\begin{itemize}
    \item Developed high-fidelity simulation environments and generated multiple datasets to support the development and performance evaluation of lunar rover navigation with crater landmarks\\
    \item Generated a high-resolution crater database of real Lunar data from the Chang’e 4 landing site.\\
    \item Raised the TRL of onboard crater detection capability from 2 to 4 by successfully demonstrating crater detection algorithms on both simulated and real Lunar data.\\
    \item Raised the TRL of onboard, global (“absolute”) localization capability from 2 to 4 by successfully demonstrating crater-based localization algorithms on both simulated and real Lunar data\\
    \item Completed software and dataset documentation (to accompany final publication of this manuscript) to enable seamless re-distribution and open-source release, for future funded efforts and wide-spread use by the community.
\end{itemize}

\subsection{Remaining Gaps, Risks and Challenges to Flight Infusion}
As mentioned in Section \ref{sec:intro}, the capability developed in this project is intended for use in lunar rovers. No such missions are currently in development, but robotic lunar science rover mission concepts were strongly recommended by the PSADS report; in particular, the Endurance-A robotic lunar science rover mission concept was recommended as the highest priority medium-sized mission for the Moon. This concept involves traverse of roughly 2,000 km in several Earth years, which requires onboard absolute position estimation; LunarNav capability would be enabling for this mission. The Lunar Terrain Vehicle (LTV) envisioned for transporting astronauts may also benefit from having an option for autonomous position estimation as developed by LunarNav. Some of the potential challenges to flight infusion falls into the following main categories:
\begin{itemize}
    \item The state of maturity of the LunarNav algorithms is at intermediate TRLs; more maturation is required before it is fully ready for infusion. In particular, the performance of stereo-based crater localization needs to be improved by adding robustness to a wide-range of operational scenarios (partial crater visibility, variable crater shapes) \\
    \item LunarNav requires stereo cameras and/or a lidar onboard the rover. For night operation, either headlights for the stereo cameras or a lidar is required. These sensors are not fully developed at present. Furthermore, the majority of the work done in LunarNav focused on day-time driving; further work needs to be done to extend this capability to night-time and operations in permanently shadowed regions. \\
    \item The computing load associated with crater-based localization is expected to be less than that required for rover obstacle avoidance. Since any autonomous lunar rover would need obstacle avoidance, it is expected that the necessary computing capability for LunarNav would be available. Nevertheless, this aspect of system architecture must be verified as being sufficient. \\
    \item Validation and verification (V\&V) of crater-based localization requires datasets with craters, either from lunar analog terrain on Earth or from high fidelity lunar terrain simulators. There is a very limited amount of suitable analog terrain available. The Cinder Lake Crater Field planned for use in the LunarNav project is the only location of any reasonable size in the U.S.; it is nevertheless fairly small for this purpose, it has not been maintained (i.e. it is partly overgrowth with vegetation), and access to it is limited to spring and fall seasons, due to snow in the winter, heat in the summer, and the possibility of nearby forest fires in the summer through early fall.
\end{itemize}

\acknowledgments
The research was carried out at the Jet Propulsion Laboratory, California Institute of Technology, under a contract with the National Aeronautics and Space Administration (80NM0018D0004).

\bibliographystyle{IEEEtran}
\bibliography{references}

\thebiography
\begin{biographywithpic}
{Shreyansh Daftry}{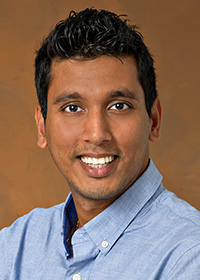}
is a Robotics Technologist at NASA Jet Propulsion Laboratory, California Institute of Technology. He received his M.S. degree in Robotics from the Robotics Institute, Carnegie Mellon University, and his B.S. degree in Electronics and Communication Engineering. His research interest lies is at intersection of space technology and autonomous robotic systems, with an emphasis on machine learning applications to perception, planning and decision making. At JPL, he is the Group Leader of the Perception Systems group, is working on the Mars Sample Recovery Helicopter mission, and has led/contributed to several technology development for autonomous navigation of ground, airborne and subterranean robots.
\end{biographywithpic}

\begin{biographywithpic}
{Zhanlin Chen}{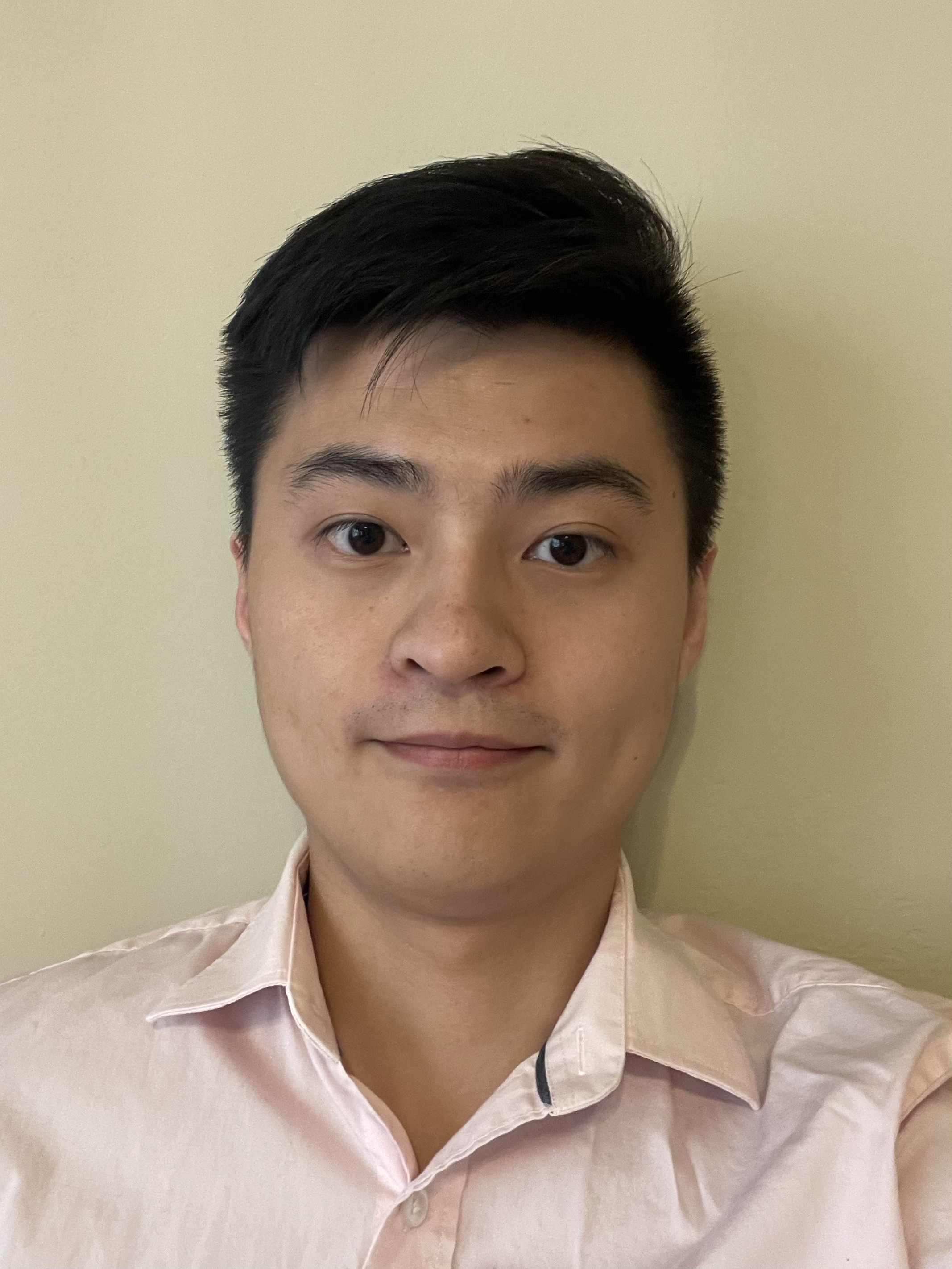}
is a Robotics Engineering Intern within the 347J Perception Systems section at the Jet Propulsion Laboratory. His research interests span perception, statistics, and machine learning. He received his B.S. and M.S. degrees from Yale University in Computer Science and Statistics and Data Science. 
\end{biographywithpic}

\begin{biographywithpic}
{Yang Cheng}{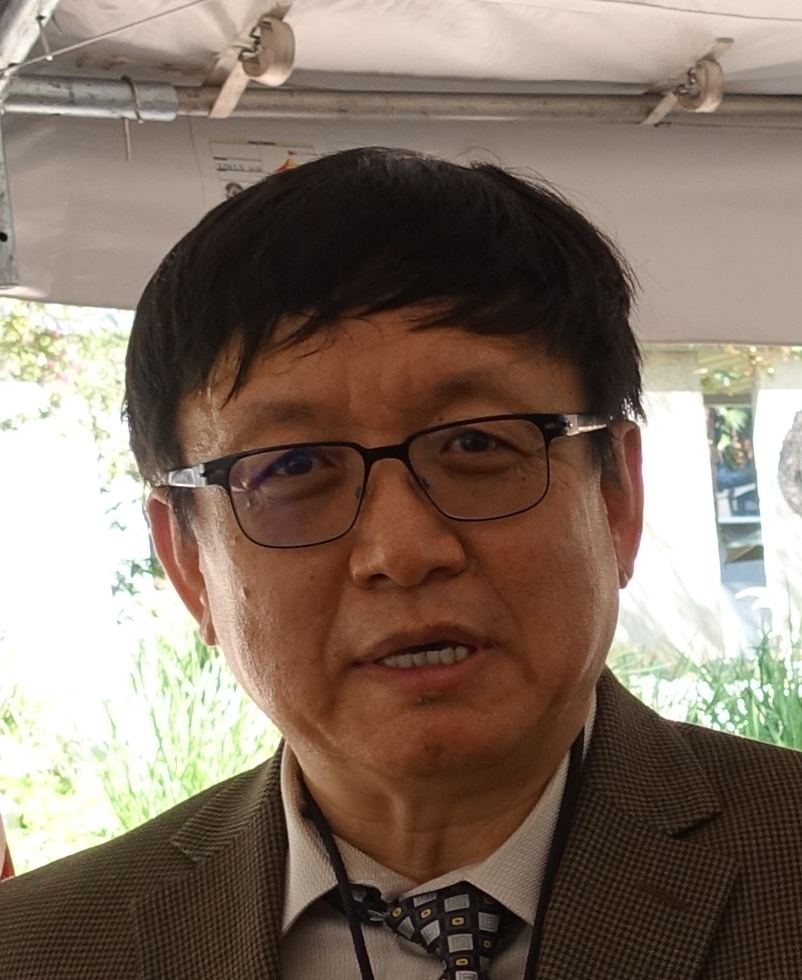}
is a principal member of the Aerial System Perception Group at JPL.  Dr. Cheng is a leading expert in the area of optical spacecraft navigation, terrain relative navigation, surface robotic perception and cartography. He has made significant technical contributions in technology advancement in the area of structure from motion, 3D surface reconstruction, surface hazard detection and mapping for spacecraft landing site selection, stereo matching and sub-pixel interpolation, map projection. Dr. Cheng is the key  developer for MER descent image motion estimation system (DIMES) and Mars2020 lander vision system (LVS) and  the first ever onboard reference map for Mars2020 LVS.
\end{biographywithpic}

\begin{biographywithpic}
{Scott Tepsuporn}{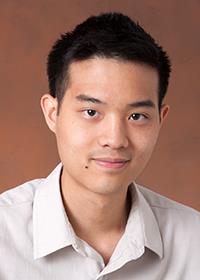}
Scott Tepsuporn is a Robotics Electrical Engineer within the Mobility and Robotic Systems section at the Jet Propulsion Laboratory where he is involved in the design and implementation of various motion planning, computer vision, and mission autonomy work. He received his B.S. degrees from the University of Virginia in Computer Science and Electrical Engineering.
\end{biographywithpic}

\begin{biographywithpic}
{Brian Coltin}{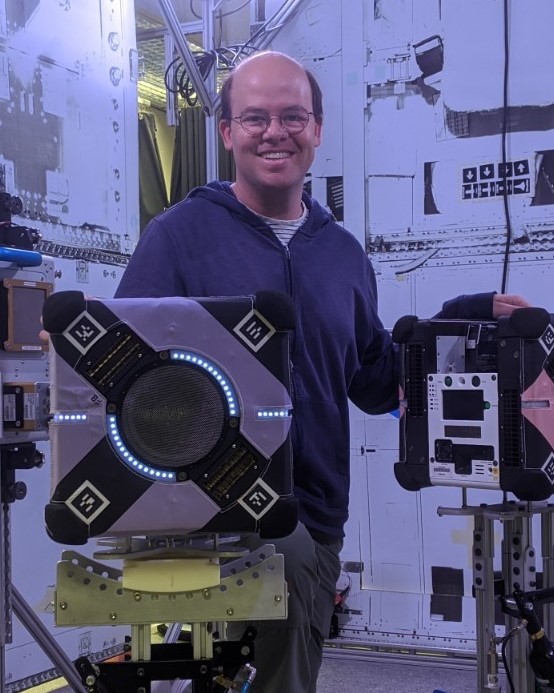}
is a roboticist with the Intelligent Robotics Group at NASA Ames. He currently leads software development for the Astrobee robots. His research interests include localization, planning, scheduling, and computer vision. He received his B.S. in computer science (2009) and Ph.D. in robotics (2014) from Carnegie Mellon University.
\end{biographywithpic}

\begin{biographywithpic}
{Ussama Naal}{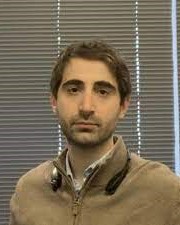}
 received his B.S. degree in Informatics Engineering in 2007 from University of Aleppo and a M.S. in Electrical and Computer Engineering from the University of Oklahoma in 2011. His expertise lies within areas of image processing, data visualisation, computer graphics and simulation. Ussama joined NASA Ames in 2020 and contributed to multiple projects in the field of robotics and simulation.
\end{biographywithpic}

\begin{biographywithpic}
{Lanssie Mingyue Ma}{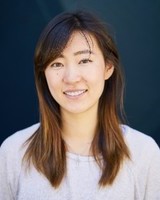}
is a software engineer within the Intelligent Robotics Group at NASA Ames. She is involved in the UI/UX for fleet management, path planning for rovers, and Augmented and Virtual Reality scenarios in far-future lunar operations for crew. She received her B.A. in Computer Science from the University of California, Berkeley, and her M.S. and Ph.D. from Georgia Tech in human-robot teaming.
\end{biographywithpic}

\begin{biographywithpic}
{Shehryar Khattak}{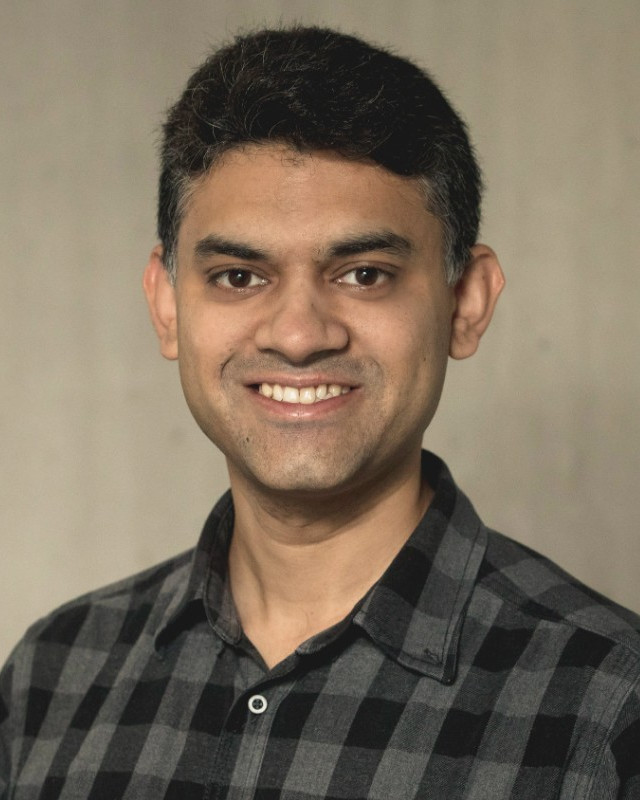}
is a Robotics Technologist within the Perception Systems Group at the NASA Jet Propulsion Laboratory. Before joining JPL, he was a post-doctoral researcher at ETH Zurich and received his Ph.D. (2019) and MS (2017) in Computer Science from the University of Nevada, Reno. He also holds an MS in Aerospace Engineering from KAIST (2012) and a BS in Mechanical Engineering from GIKI (2009). Previously, he worked at Samsung Electronics, South Korea (2012-2015) on visual inspection of NAND memory devices. At JPL, his work focuses on enabling resilient robot navigation in difficult environments through multi-sensor information fusion.
\end{biographywithpic}

\begin{biographywithpic}
{Matthew Deans}{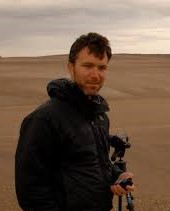}
received his PhD in Robotics from Carnegie Mellon University in 2005. His research interests include robotic mapping, localization, navigation, machine vision in unstructured environments, remote operation of rovers, and field robotics especially in terrestrial planetary analogs. On Resource Prospector he served as the lead for Rover Navigation.
\end{biographywithpic}

\begin{biographywithpic}
{Larry Matthies}{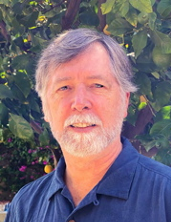}
is the technology coordinator in the Mars Exploration Program Office at JPL. He received B.S., M. Math, and PhD degrees in Computer Science
from the University of Regina (1979), University of Waterloo (1981), and Carnegie Mellon University (1989). He has been at JPL for more than
33 years, where he supervised the Computer Vision group for 21 years. He has led technology development in perception systems for
autonomous navigation of robotic vehicles for land, sea, air, and space applications on Earth and in planetary exploration, including development of computer vision algorithms for Mars rovers, landers, and helicopters. He is a Fellow of the IEEE and a member of the editorial boards of Autonomous Robots and the Journal of Field Robotics.
\end{biographywithpic}

\end{document}